\documentclass[journal, 10pt, twocolumn]{IEEEtran}

\usepackage{graphicx} 
\usepackage{subfigure}
\usepackage{caption}
\usepackage{amsmath}
\usepackage{mathrsfs}
\usepackage{bm}
\usepackage{comment}
\usepackage{multicol}
\usepackage{setspace}
\usepackage{pdfpages}
\usepackage{setspace}
\usepackage[numbers,sort&compress]{natbib}
\usepackage{multirow}
\usepackage{array}
\usepackage[ruled]{algorithm2e}
\usepackage{pifont}
\usepackage[switch]{lineno} 
\usepackage{booktabs}
\usepackage{longtable}
\usepackage{tabularx}
\usepackage{fancyhdr} 
\usepackage{fontsize} 
\usepackage{url}

\usepackage{tikz}
\newcommand*{\circled}[1]{\lower.7ex\hbox{\tikz\draw (0pt, 0pt)%
    circle (.5em) node {\makebox[1em][c]{\small #1}};}}

\title{\LARGE {AuthSim: Towards Authentic and Effective Safety-critical Scenario Generation for Autonomous Driving Tests}
}

\author{Yukuan Yang$^{1}$, Xucheng Lu$^{2}$, Zhili Zhang$^{1}$, Zepeng Wu$^{1,3}$, Guoqi Li$^{4}$, Lingzhong Meng$^{1}$, Yunzhi Xue$^{1}$ \\
\vspace{5pt}

\small $^1$Institute of Software, Chinese Academy of Sciences, Beijing 100190, China. \\
\small $^2$School of Science, Beijing University of Posts and Telecommunications, Beijing 100876, China. \\
\small $^3$ University of Chinese Academy of Sciences, Beijing 101408, China. \\
\small $^4$ Institute of Automation, Chinese Academy of Sciences, Beijing 100190, China.

\thanks{yangyukuan@iscas.ac.cn (Y. Yang)}
}

\begin{document}
\maketitle
\thispagestyle{fancy}
\pagestyle{fancy} 
\fancyhf{} 
\cfoot{\thepage} 
\renewcommand{\headrulewidth}{0pt} 
\renewcommand{\footrulewidth}{0pt} 
\cfoot{\fontsize{9}{11}\selectfont\thepage} 

\newcommand{\hy}[1]{\textcolor{red}{(hy: #1)}}

\begin{abstract}
    Generating adversarial safety-critical scenarios is a pivotal method for testing autonomous driving systems, as it identifies potential weaknesses and enhances system robustness and reliability. However, existing approaches predominantly emphasize unrestricted collision scenarios, prompting non-player character (NPC) vehicles to attack the ego vehicle indiscriminately. These works overlook these scenarios' authenticity, rationality, and relevance, resulting in numerous extreme, contrived, and largely unrealistic collision events involving aggressive NPC vehicles. To rectify this issue, we propose a three-layer relative safety region model, which partitions the area based on danger levels and increases the likelihood of NPC vehicles entering relative boundary regions. This model directs NPC vehicles to engage in adversarial actions within relatively safe boundary regions, thereby augmenting the scenarios' authenticity. We introduce AuthSim, a comprehensive platform for generating authentic and effective safety-critical scenarios by integrating the three-layer relative safety region model with reinforcement learning. To our knowledge, this is the first attempt to address the authenticity and effectiveness of autonomous driving system test scenarios comprehensively. Extensive experiments demonstrate that AuthSim outperforms existing methods in generating effective safety-critical scenarios. Notably, AuthSim achieves a 5.25\% improvement in average cut-in distance and a 27.12\% enhancement in average collision interval time, while maintaining higher efficiency in generating effective safety-critical scenarios compared to existing methods. This underscores its significant advantage in producing authentic scenarios over current methodologies.
\end{abstract}

\section{Introduction}
\label{sec:intro}


Over the past decade, autonomous driving vehicles \cite{yurtsever2020survey,chen2021interpretable,singh2022road} have made significant strides, largely due to the rapid development and application of machine learning \cite{jordan2015machine,lecun2015deep}. These vehicles are anticipated to become the cornerstone of future transportation systems, offering enhancements in safety, efficiency, intelligence, and environmental protection \cite{lv2022impacts,jabbar2022blockchain}. Despite this progress, autonomous driving vehicles face substantial security threats due to the opaque nature and uncertain boundaries of intelligent systems \cite{ju2022survey}. These safety concerns present a significant obstacle to the widespread adoption of autonomous driving technology.


In this context, testing the safety of autonomous driving vehicles becomes crucial \cite{wang2020safety}. On one hand, tests can quickly identify safety issues, facilitating rapid updates and iterations of autonomous driving technology. On the other hand, testing and evaluating the maturity and safety of these vehicles are essential for determining their suitability for large-scale road deployment. Unlike the mechanical and dynamic tests conducted on traditional non-intelligent vehicles, the tests for autonomous driving vehicles focus primarily on their intelligent driving capabilities \cite{adler2016safety}.


Autonomous driving vehicle tests are typically conducted in two primary ways: test sites and simulations. Many countries and companies are investing in the construction of autonomous driving test sites to facilitate research and development. Notable examples include Mcity \cite{dong2019mcity}, Waymo's Castle \cite{liu2021towards}, GoMentum Station \cite{cosgun2017towards}, the Singapore Autonomous Vehicle Initiative \cite{tan2021adaptive}, and CARIAD \cite{zhang2022aerial}. Although these test sites offer a controlled environment for researchers and developers to test and refine autonomous vehicle technology, they have several potential limitations. These include a lack of real-world scalability, limited real-world experience, high construction and maintenance costs, safety concerns, and possible biases in testing scenarios \cite{gambi2019automatically}. As an alternative, simulation tests address these drawbacks and provide a valuable tool for improving and refining autonomous driving vehicles \cite{dosovitskiy2017carla,rong2020lgsvl}. Simulation tests use virtual environments and computer simulations to evaluate and validate the performance and safety of autonomous driving systems. These simulations offer a controlled and repeatable environment, enabling researchers and developers to reproduce scenarios, customize tests, validate scalability, and identify and address issues more efficiently.


Generating safety-critical scenarios is essential for autonomous vehicle simulation tests \cite{ding2020learning}. However, due to the low collision rate in real-world driving environments, identifying and collecting these scenarios can be challenging \cite{song2023critical,makansi2021exposing}. Consequently, it is crucial to artificially generate safety-critical scenarios. There are three primary methods for generating these scenarios: data-driven methods \cite{knies2020data,ding2020cmts,suo2021trafficsim,chen2021geosim}, knowledge-based methods \cite{fremont2020formal,wang2021commonroad,mcduff2022causalcity,giamattei2024causality}, and adversarial generation methods \cite{feng2021intelligent,chen2021adversarial,koren2021finding,rempe2022generating, biagiola2024boundary}. Data-driven methods sample real-world datasets, estimate scenario distributions and learn a generative model to create new scenarios. Knowledge-based methods incorporate domain knowledge, such as traffic rules and physical laws, to efficiently generate scenarios. Adversarial generation methods create safety-critical scenarios by introducing aggressive Non-Player Character (NPC) vehicles into the driving environment. Compared to data-driven methods, which require extensive data collection, and knowledge-based methods, which rely heavily on expert experience, adversarial generation methods offer a cost-effective, efficient, and reliable means of identifying corner cases.


Existing adversarial generation methods are commonly used to create safety-critical scenarios. These methods establish an objective function for scenario criticality evaluation and iteratively optimize the parameters of NPC vehicles. However, current research faces a major challenge: generating effective safety-critical collision scenarios that maintain NPC vehicle adversarial rationality while ensuring the efficiency of scenario generation.

\begin{figure}[!htpb]
	\centering
\includegraphics[width=0.48\textwidth]{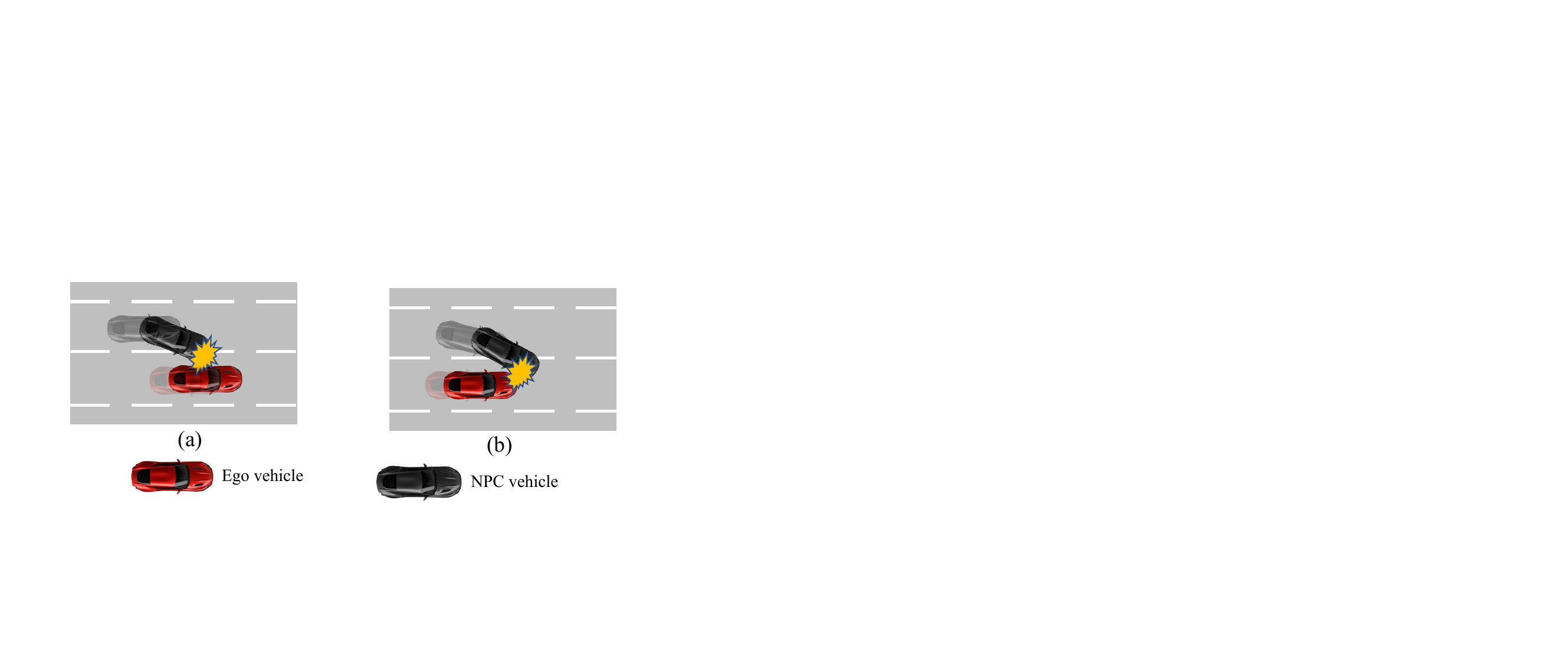} 
	\caption{Examples of extreme and unreasonable collision scenarios for autonomous driving tests. (a) NPC side attacks; (b) NPC dangerous lane changes.}
	\label{fig:extre_collision}
\end{figure}


The adversarial rationality issue arises from the design of the objective function, which encourages NPC vehicles to engage in any form of attack on the ego vehicle \cite{li2020av,tian2022mosat}. This leads to extreme and deliberate behaviors, such as NPC vehicles actively colliding with ego vehicles from the side (Figure \ref{fig:extre_collision} (a)) or intentionally changing lanes within a very short distance (Figure \ref{fig:extre_collision} (b)). These scenarios often violate physical limits, making it impossible for the ego vehicle to avoid a collision regardless of its actions. From a responsibility allocation perspective, even in the event of a collision, the majority of responsibility typically lies with NPC vehicles rather than ego vehicles. These scenarios are widely present in current methods. On the one hand, while many collision scenarios are generated, a significant portion of these scenarios is extreme and ineffective, resulting in low efficiency in producing effective safety-critical scenarios. On the other hand, the rationality of adversarial scenarios is very low, and such scenarios are rarely observed in actual natural driving conditions, thus offering little value for autonomous driving tests.

In this research, we aim to address the issues of NPC vehicles' adversarial rationality and scenario generation efficiency in the context of generating adversarial scenarios for autonomous driving tests. To overcome these challenges, we introduce a novel method for evaluating scenario criticality and an innovative optimization approach. Through extensive experimentation, we demonstrate the efficacy of our proposed methods. Our contributions can be summarized as follows:
\begin{itemize}
\item We propose a novel method for evaluating scenario criticality that incorporates a real-time three-layer relative safety region model to ensure the authenticity and naturalness of adversarial scenarios. Based on the different reaction times available to the ego vehicle, we model the region in front of the ego vehicle in real-time as the relative danger region, the relative boundary region, and the relative safety region. Scenario criticality is calculated using the residence time, occupied area, and empirical probability values of NPC vehicles entering these different regions.




\item We develop a comprehensive platform called AuthSim for authentic and effective generation of safety-critical scenarios, which combines the three-layer relative safety region model with reinforcement learning. We design the state and behavior space of NPC vehicles and dynamically control them using reinforcement learning models. The three-layer relative safety zone model is trained as a reward function to generate adversarial scenarios.


\item Extensive experiments have been conducted to verify the efficiency and authenticity of the proposed AuthSim framework in generating safety-critical scenarios. Compared to classic methods such as Naturalistic and Adversial Driving Environment (NADE), Time to Collision (TTC), Time to Brake (TTB), and Deceleration Rate to Avoid a Crash (DRAC), AuthSim demonstrates marginally higher efficiency in generating effective safety-critical scenarios while achieving significantly higher authenticity. AuthSim produces scenarios with notably greater cut-in distances and longer collision time intervals, indicating substantial authenticity improvements in safety-critical scenario generation.



\end{itemize}


The paper is organized as follows: Section \ref{sec:related_works} provides an overview of recent works related to safety-critical scenario generation. Section \ref{sec:methods} details the proposed three-layer relative safety region model and the authentic scenario generation framework, AuthSim. Section \ref{sec:exp_results} presents the experimental results and analysis of AuthSim compared to existing methods. Finally, Section \ref{sec:conclusion} summarizes the paper and offers concluding remarks.

\section{Related Works}
\label{sec:related_works}

\textbf{Adversarial safety-critical scenario generation:} Adversarial methods efficiently and cost-effectively generate safety-critical test scenarios for autonomous driving, unlike data-driven and knowledge-driven methods. However, current research often focuses on unrestricted collision scenarios, neglecting their authenticity, rationality, and significance for autonomous driving tests. NADE \cite{feng2021intelligent} establishes a naturalistic and adversarial driving environment for generating safety-critical scenarios by collecting driving data from natural traffic and employing importance sampling on specific vehicles to interact with ego vehicles, thereby enhancing the efficiency of safety-critical scenario generation. AV-FUZZER \cite{li2020av} generates autonomous driving safety-critical scenarios by perturbing NPC vehicle maneuvers and using a genetic algorithm to optimize these perturbations and define trajectories. MOSAT \cite{tian2022mosat} constructs diverse and adversarial driving environments to expose autonomous driving safety violations by using atomic driving maneuvers, motif patterns, and a multi-objective genetic algorithm. AdvSim \cite{wang2021advsim} generates safety-critical scenarios for LiDAR-based autonomous systems by modifying actors' trajectories in a physically plausible way and updating the corresponding LiDAR sensor data. D2RL \cite{feng2023dense} trains NPC vehicles using dense deep-reinforcement learning by refining Markov decision processes, removing non-safety-critical states, and reconnecting critical ones to prioritize safety-critical scenarios. SDC-Prioritizer \cite{birchler2023single} introduces two evolutionary methods, SO-SDC-Prioritizer and MO-SDC-Prioritizer, which utilize single-objective and multi-objective genetic algorithms to prioritize test scenarios based on diversity metrics without past execution results, thereby balancing cost-effectiveness and test diversity. Similar works can also be found in \cite{abeysirigoonawardena2019generating,li2021scegene,tian2022generating}. These studies mostly focus on enhancing the collision rate of generated scenarios. However, they often overlook the authenticity of these scenarios, the allocation of accident responsibilities, and their relevance to autonomous driving tests. Recently, there has been increasing concern regarding the authenticity of testing scenarios for autonomous driving vehicles. STRIVE \cite{rempe2022generating} introduces an innovative method utilizing a graph-based conditional variational autoencoder (CVAE) to generate challenging scenarios for autonomous driving planners automatically. However, this work is only focused on the planner module of an autonomous driving system, with no involvement in real scenario generation for the entire autonomous driving system.

\section{Methods}
\label{sec:methods}


We aim to generate authentic safety-critical scenarios for autonomous driving tests with high efficiency. Specifically, NPC vehicles take reasonable adversarial behaviors toward the Ego vehicle leading to collisions  with the Ego vehicle assuming primary responsibility. The efficiency is defined as generating as many valid adversarial scenarios as possible within the same number of tests. In this section, we will first elucidate the classic and widely utilized scenario criticality objectives for adversarial scenario generation, including Time to Collision (TTC), Time to Brake (TTB), and Deceleration Rate to Avoid a Crash (DRAC), which also serve as baselines in Section \ref{sec:exp_results}. Subsequently, we will expound upon the real-time three-layer relative safety region model. Finally, we will present the methodology for authentic safety-critical scenario generation, AuthSim, which integrates the three-layer relative safety region model with reinforcement learning.


\subsection{Traditional scenario criticality objectives}
\label{sec:sce_cri_obj}


Prior to our work, three traditional criticality objectives have been extensively employed in scenario generation: TTC \cite{hayward1972near}, TTB \cite{hillenbrand2005situation}, and DRAC \cite{li2017crash}. These metrics have served as foundational benchmarks in the field of adversarial scenario generation for autonomous driving systems.



Time to Collision (TTC)  is a widely recognized safety-related metric that quantifies the time remaining until a collision between the ego vehicle and NPC vehicles is expected to occur, given their current velocities and trajectories. Formally, TTC is defined as
\begin{equation}
\centering
TTC=\frac{\Delta p}{v_{i}-v_{i-1}},
\label{equ:ttc}
\end{equation}
where $\Delta p$ represents the distance between the following vehicle and the leading vehicle, $v_{i}$ denotes the speed of the following vehicle, and $v_{i-1}$ denotes the speed of the leading vehicle. This metric is instrumental in assessing the immediacy of potential collisions and is crucial for the development and evaluation of autonomous driving safety systems.


Time to Brake (TTB) represents the time required for the following vehicle to decelerate at maximum deceleration to match the speed of the lead vehicle. Mathematically, it can be expressed as


\begin{equation}
\centering
TTB=TTC + \frac{v_{rel}}{2a_{ego,max}},
\label{equ:ttb}
\end{equation}
where $TTC$ is the time to collision metric defined in Equation \ref{equ:ttc}, $v_{rel}$ represents the relative speed between the following and leading vehicles, and $a_{ego,max}$ denotes the maximum deceleration that the ego vehicle can execute.


The Deceleration Rate to Avoid a Crash (DRAC) metric quantifies the deceleration required by the following vehicle to prevent a collision with the leading vehicle. A higher DRAC value signifies an increased risk of collision. If the DRAC metric surpasses the maximum deceleration capability of the following vehicle, deceleration alone will be inadequate to avoid a collision. The DRAC is described as

\begin{equation}
\centering
DRAC = \frac{\left(v_i - v_{i-1}\right)^2}{2\Delta p},
\label{equ:DRAC}
\end{equation}
\do{where $v_i$ is the speed of the following vehicle and $v_{i-1}$ is the speed of the leading vehicle, and $\Delta p$ is the distance between the following and the leading vehicles.}


\subsection{Three-layer relative safety region model-based scenario criticality objective}
\label{sec:saf_cri_obj}


The Responsibility-Sensitive Safety (RSS) model, as outlined in \cite{ShalevShwartz2017OnAF}, defines longitudinal and lateral safety distances that vehicles should uphold to enable prompt response, factoring in reaction times, while preventing collisions. This model prescribes specific distances and recommends appropriate maneuvers for vehicles to avoid potential hazards. Expanding upon this framework, we have broadened the concept of safety distances to encompass danger distances, boundary distances, and safety distances between vehicles. Leveraging these extended distance metrics, we propose a real-time three-layer relative safety region model.

\begin{figure}[htpb]
	\centering
	\subfigure[]{\includegraphics[width=0.48\textwidth]{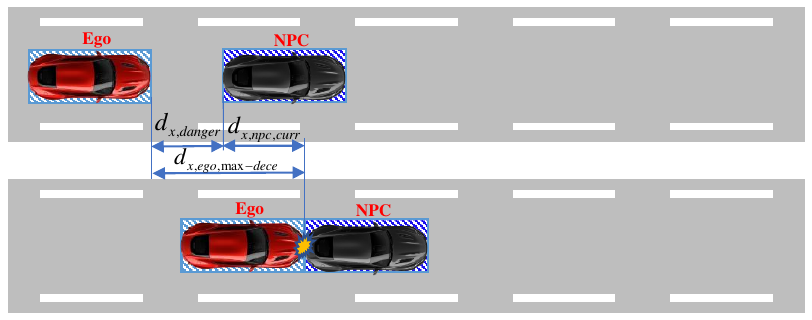} \label{fig:danger}}
        \centering
	\subfigure[]{\includegraphics[width=0.48\textwidth]{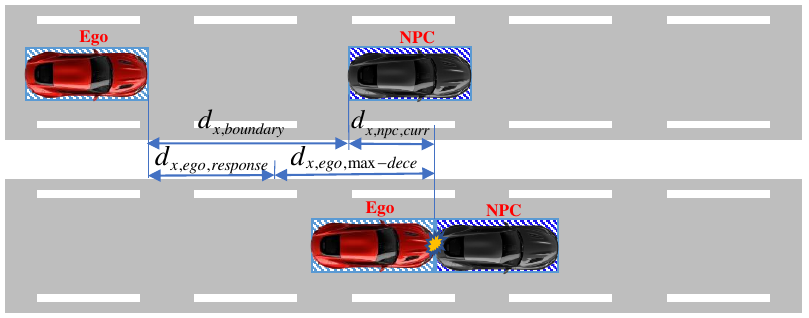} \label{fig:boundary}}
     \centering
     \subfigure[]{\includegraphics[width=0.48\textwidth]{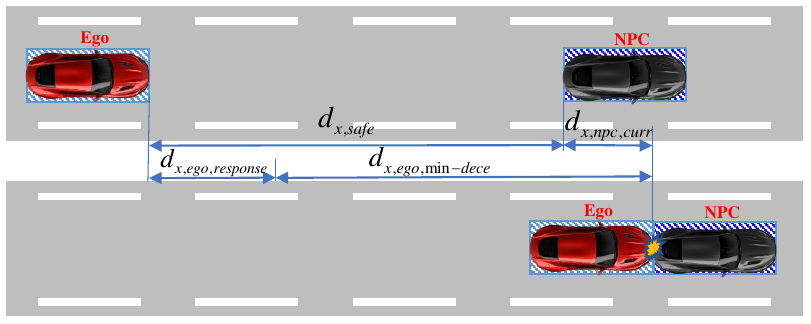} \label{fig:safe}}
	\caption{Illustration of longitudinal danger, boundary, and safety distances between the ego and NPC vehicles: (a) Danger distance; (b) Boundary distance; (c) Safety distance.}
	\label{fig:x_distance}
\end{figure}


Figure \ref{fig:danger} illustrates the real-time relative longitudinal danger distance. When an NPC vehicle enters the collision danger distance relative to the ego vehicle, a collision becomes inevitable if the NPC vehicle continues in its current state, regardless of any emergency maneuvers executed by the ego vehicle. Assuming the ego vehicle is positioned at the origin of the coordinate system, with its front facing the positive direction of the x-axis (longitudinal axis), and the NPC vehicle traveling at its current speed, a collision will occur if the distance between the vehicles is less than this critical distance, even if the ego vehicle applies maximum deceleration immediately, with zero reaction time. In this scenario, the longitudinal danger distance between the ego vehicle and the NPC vehicle along the x-axis can be calculated as
\begin{eqnarray}
\begin{split}
d_{x, danger}= &d_{x,ego,max,dece} - d_{x,npc,curr} \\
=&\frac{v_{x,ego}^2 - v_{x,npc}^2}{2 a_{x,ego,max,dece}} - v_{x,npc} \frac{v_{x,ego} - v_{x,npc}}{a_{x,ego,max,dece}},
\label{equ:x_danger}
\end{split}
\end{eqnarray}
where $d_{x,ego,max,dece}$ represents the braking distance of the ego vehicle in the x-axis direction under maximum deceleration $a_{x, ego,max,dece}$, while $d_{x,npc,curr}$ is the distance in the x-axis direction that the NPC vehicle travels within the ego braking time at the current speed $v_{x,npc}$. $v_{x,ego}$ refers to the speed of the ego vehicle in the x-axis direction.


Similarly, we establish a real-time model for the relative longitudinal boundary distance, incorporating reaction time and accident avoidance maneuvers. Figure \ref{fig:boundary} illustrates the concept of the boundary distance, defined as the distance at which the ego vehicle, after a specified reaction time, engages in emergency braking at maximum deceleration, while the NPC vehicle continues traveling at its current speed, thereby preventing a collision. During the reaction time, the ego vehicle accelerates at maximum acceleration without perceiving any danger. Suppose the distance between the ego and NPC vehicles is less than this boundary distance and the NPC vehicle maintains a constant speed as per its current state. In that case, the ego vehicle must initiate emergency braking within the specified response time to avert a collision. Thus, we define the relative longitudinal boundary distance between the ego and NPC vehicles in the x-axis direction as
\small
\begin{eqnarray}
\begin{split}
&d_{x,boundary}= d_{x,ego,response}+d_{x,ego,max,dece} - d_{x,npc,curr} \\
&= v_{x,ego} \rho+\frac{1}{2} a_{ x,ego, max, accel} \rho^2 \\
&+\frac{\left(v_{x,ego} +a_{x,ego,max,accel} \rho\right)^2 - v_{x,npc}^2}{2 a_{x,ego,max,dece}}\\ 
& -v_{x,npc}\left(\rho+\frac{v_{x,ego}+a_{ x,ego, max, accel} \rho - v_{x,npc}}{a_{x, ego, max , dece}}\right), \\
\end{split}
\label{equ:x_boundary}
\end{eqnarray}
\normalsize
where $d_{x,ego,response}$ and $a_{x, ego, max, accel}$ are the response distance and maximum acceleration of the ego vehicle, respectively. $\rho$ signifies the response time.


Finally, the real-time relative longitudinal safety distance is illustrated in Figure \ref{fig:safe}. This model stipulates that the ego vehicle begins reacting after a specified period and subsequently applies brakes at the minimum deceleration rate. Meanwhile, the NPC vehicle continues to travel at a consistent speed, ensuring a collision-free scenario. Suppose the distance between the ego and NPC vehicles exceeds this threshold, and the NPC vehicle maintains its current speed based on prevailing conditions. In that case, a collision will be avoided as long as the ego vehicle initiates braking after a certain duration. Thus, the longitudinal safety distance between the ego and NPC vehicles can be determined as follows:
\begin{eqnarray}
\begin{split}
&d_{x,safety}= d_{x,ego,response}+d_{x,ego,min,dece} - d_{x,npc,curr} \\
&= v_{x,ego} \rho+\frac{1}{2} a_{ x,ego, max, accel} \rho^2 \\
&+\frac{\left(v_{x,ego} +a_{x,ego,max,accel} \rho\right)^2 - v_{x,npc}^2}{2 a_{x,ego,min,dece}}\\ 
& -v_{x,npc}\left(\rho+\frac{v_{x,ego}+a_{ x,ego, max, accel} \rho - v_{x,npc}}{a_{x, ego, min , dece}}\right), \\
\end{split}
\label{equ:x_safe}
\end{eqnarray}
where $d_{x,ego,min,dece}$ represents the distance traveled by the ego vehicle after the response time $\rho$ with minimum deceleration $a_{x, ego, min, dece}$.



Considering that the y and x directions are consistent, the real-time relative lateral danger, boundary, and safety distances in the y-axis direction are

\begin{eqnarray}
\begin{split}
&d_{y, danger}= d_{y,ego,max,dece} - d_{y,npc,curr}, \\
&d_{y,boundary}= d_{y,ego,response}+d_{y,ego,max,dece} \\
&~~~~~~~~~~~~~~~~-d_{y,npc,curr}, \\
&d_{y,safety}= d_{y,ego,response}+d_{y,ego,min,dece} - d_{y,npc,curr}, \\
\end{split}
\label{equ:y_distance}
\end{eqnarray}
where $d_{y, danger}$, $d_{y, boundary}$, and $d_{y, safety}$ represent the real-time relative lateral distances in the y-axis direction, corresponding to danger, boundary, and safety, respectively. These distances are calculated analogously to the distances in the x-axis direction.



\begin{figure}[!htpb]
	\centering
\includegraphics[width=0.48\textwidth]{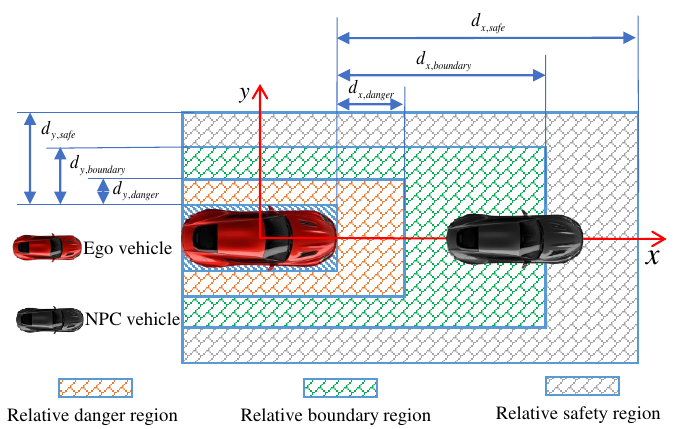} 
	\caption{The three-layer relative safety region model.}
	\label{fig:region_model}
\end{figure}

By integrating the various safety distances in both the x-axis and y-axis directions, we develop a real-time three-layer relative safety region model. As illustrated in Figure \ref{fig:region_model}, this model comprises three distinct regions: the relative danger region, the relative boundary region, and the relative safety region.

\begin{figure}[!htpb]
	\centering
\includegraphics[width=0.48\textwidth]{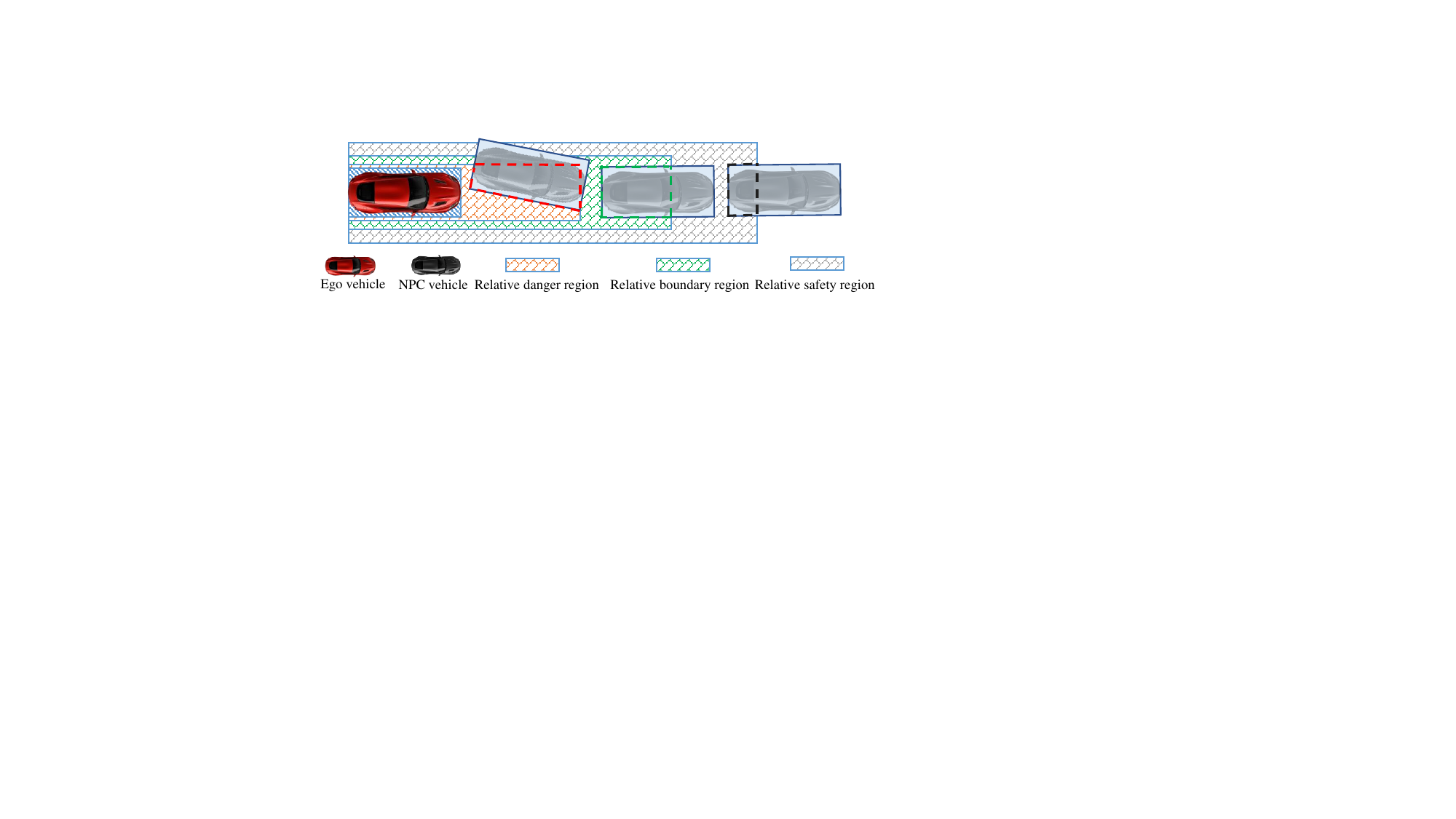} 
	\caption{Different overlapping area and staying time of the NPC vehicle in different relative regions.}
	\label{fig:region_area}
\end{figure}

After implementing the real-time three-layer relative safety region model, we integrate it into the criticality model for scenario evaluation. As depicted in Figure \ref{fig:region_area}, the NPC vehicle traverses various regions—danger, boundary, and safety—during its maneuvers. These regions overlap with different areas, and the NPC vehicle spends varying durations within them. It is evident that the larger the overlapping area between the NPC vehicle and the relative danger region, and the longer the NPC vehicle remains in this region, the greater the threat it poses to the ego vehicle. Conversely, when the NPC vehicle is located in a relative safety region, it poses fewer threats to the ego vehicle.

In addition to the factors mentioned above, such as overlapping area and staying time, we also need to consider the probability of the NPC vehicle appearing in different regions to closely resemble natural and realistic driving scenarios. If our focus is solely on generating collision scenarios, then we only need to optimize the overlapping area and staying time of the NPC vehicle in the relative danger region to be sufficiently large. However, this approach would only generate dangerous and extreme collision scenarios, such as lane changes at extremely close distances that even human drivers would struggle to avoid. These scenarios are relatively rare in real-world driving situations and have limited significance for testing ego vehicles. Conversely, if we prioritize lane changes within the relative safety region, the generated scenarios would pose little threat to the ego vehicle and would hardly challenge autonomous vehicles. The ideal scenarios are boundary scenarios, where the NPC vehicle poses a safety threat to the ego vehicle while exhibiting certain driving rationality. These scenarios occur more frequently in natural driving situations and can pose a significant threat to autonomous driving vehicles, forcing them to react and make correct decisions to avoid collisions.

\begin{figure}[htpb]
	\centering
\includegraphics[width=0.45\textwidth]{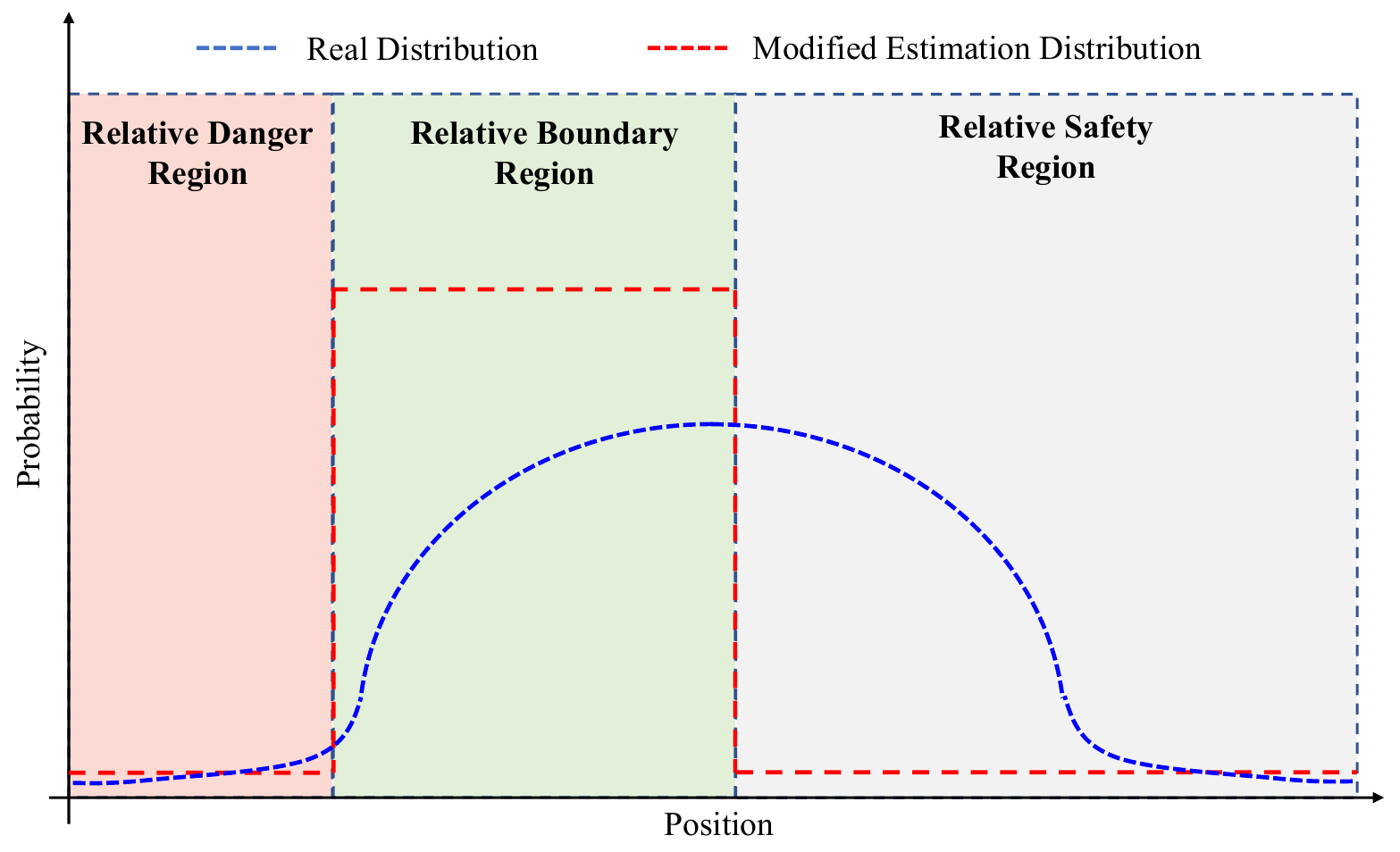} 
	\caption{Real and modified estimation distributions of NPC vehicles appearing in different regions.}
\label{fig:modified_region_distribution}
\end{figure}

To generate optimal test scenarios that consider both the rationality of NPC vehicle behavior and the authenticity of collisions, we propose incorporating the probability of NPC vehicles appearing in different regions into the assessment of scenario criticality. Empirically, during the driving process, the NPC vehicle is more likely to appear at the relative boundary and safety regions between the ego vehicle and the NPC vehicle, while the probability of it appearing in the high-risk relative danger region is extremely low. As a result, the actual distribution of NPC vehicles appearing in different regions can be approximated as a bell-shaped curve (represented by the blue dashed line in Figure \ref{fig:modified_region_distribution}). Using the accurate real distribution would undoubtedly generate scenarios that closely resemble real-world driving environments. However, obtaining such fine-grained data, including real-time information such as the speed of NPC vehicles and ego vehicles to calculate the size of different regions, is challenging. Additionally, existing traffic datasets rarely provide such detailed data. Acquiring the real distribution of NPC vehicles in different regions would require a large amount of traffic data, which is both labor-intensive and expensive. Commonly, it is worth noting that NPC vehicles are more likely to appear in the relative safety region. If scenarios primarily consist of NPC vehicles appearing in the relative safety region, the ego vehicle would face fewer threats and make fewer mistakes. Considering these limitations, it is necessary to estimate and modify the distributions of NPC vehicles appearing in different regions cost-effectively and efficiently for generating safety-critical scenarios. In this approach, we simplify the probability of NPC vehicles appearing in different regions by assuming a uniform distribution. Furthermore, we intentionally reduce the likelihood of NPC vehicles appearing in the relative safety region while increasing the probability of their presence in the relative boundary region. This strategic adjustment is designed to elevate the threat level posed to the ego vehicle (shown as the red dashed line in Figure \ref{fig:modified_region_distribution}). By employing this simplified and adjusted distribution, we can generate safety-critical scenarios while addressing the challenges associated with obtaining precise real distribution data and providing low-level threats.

\begin{figure}[!htpb]
	\centering
\includegraphics[width=0.48\textwidth]{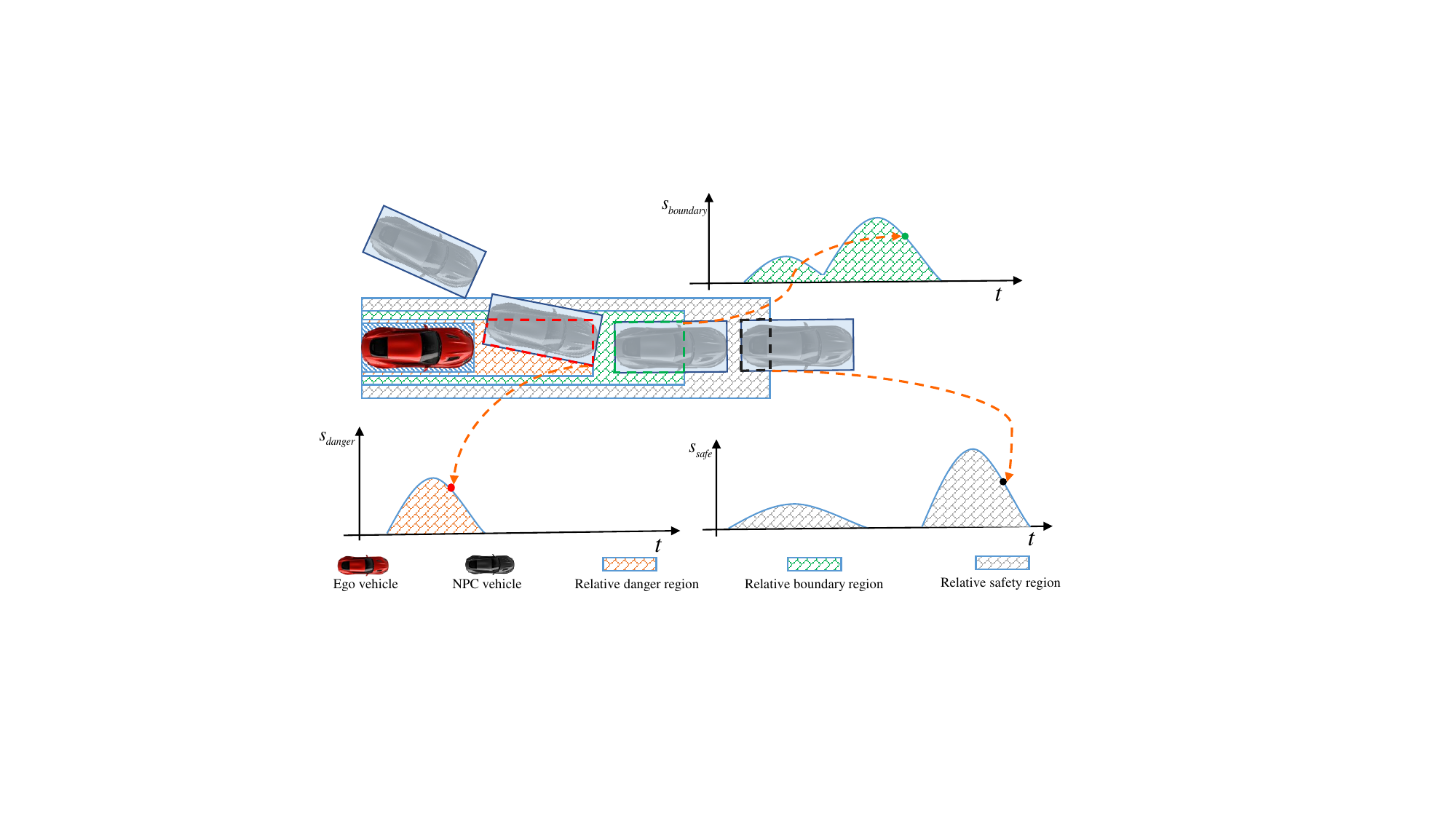} 
	\caption{Overlapping area curves over the whole driving process.}
\label{fig:area}
\end{figure}


By utilizing the real-time three-layer relative safety region model and adjusting the estimation distribution of NPC vehicles across various regions, it becomes feasible to establish the objective of scenario criticality. Figure \ref{fig:area} provides a visual representation of an NPC vehicle, depicted as a rectangle, traversing each region and remaining there for a specific duration. Throughout the entire driving process, the overlapping area between the NPC rectangle and different regions results in a temporal curve. This integration of overlapping areas between vehicles and regions in the temporal dimension, along with the probability of NPC vehicles appearing in different regions and the collision outcomes of the entire scenario, are combined to evaluate safety criticality. The real-time safety criticality objective for scenario generation is defined as follows:
\begin{eqnarray}
\begin{split}
J(\theta)= &\sum_{j=1}^T\sum_{i=1}^N\bigg( P_{danger}S_{danger}^{i,j}\\
&+P_{boundary}S_{boundary}^{i,j}+P_{safety}S_{safety}^{i,j} \bigg),
\end{split}
\label{equ:safety_criticality}
\end{eqnarray}
where $T$ denotes the time steps of the scenario and $N$ represents the number of NPC vehicles, $\theta$ is the parameter that determines the actions of the NPC vehicles and the object to be optimized. The probabilities $P_{danger}$, $P_{boundary}$, and $P_{safety}$ indicate the likelihood of an NPC vehicle entering the regions of relative danger, boundary, and safety, respectively. The variables $S_{danger}^{i,j}$, $S_{boundary}^{i,j}$, and $S_{safety}^{i,j}$ represent the real-time overlapping areas between the $i$-th NPC vehicle and the regions of relative danger, boundary, and safety at the time step $j$.

The scenario criticality objective based three-layer relative safety region model is now established. Principally, by partitioning the area according to the degree of danger and increasing the probability of NPC vehicles entering the relative boundary region, this approach enhances the rationality of NPC vehicles' interactions with the ego vehicle. Consequently, it significantly reduces the probability of extreme and intentional collisions.




\subsection{Safety-critical scenario generation}
\begin{figure*}[htpb]
	\centering
\includegraphics[width=1.0\textwidth]{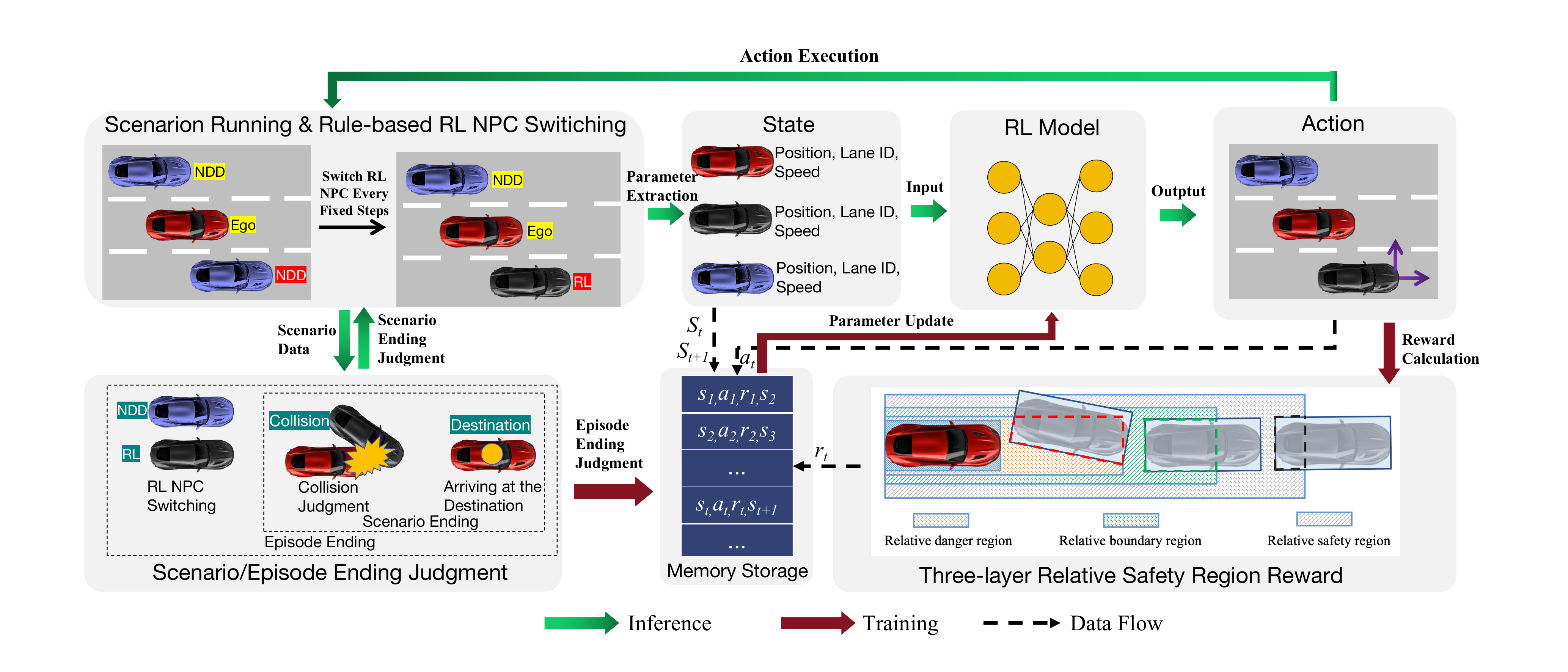} 
	\caption{Workflow of the safety-critical scenario generation framework AuthSim.}
	\label{fig:workflow}
\end{figure*}

The primary focus of generating safety-critical scenarios is to enhance and optimize the maneuvering strategies of NPC vehicles so that the attacking NPC can make authentic and rational actions when interacting with the ego vehicle. After obtaining the safety criticality objective defined in Equation \ref{equ:safety_criticality} that can guide vehicles to engage in more rational attack behavior, it's time to train the NPC vehicle's maneuver policy and generate specific safety-critical scenarios. 

Initially, we integrate the ego vehicle into the simulation system by defining its start and end points, selecting the road network, setting the scenario duration, determining the number of NDD vehicles, and randomly initializing all vehicle positions. The NDD vehicle behaviors by collecting data from natural driving scenarios, aiming to mimic human driving behavior under natural conditions closely \cite{feng2021intelligent}.

After scenario initialization, the whole workflow of the proposed safety-critical framework AuthSim is shown in Figure \ref{fig:workflow}. The safety-critical scenario generation comprises several distinct processes. Within the scenario execution and rule-based reinforcement learning (RL) NPC switching phase, the simulation system employs the RL model to retrieve vehicle action outputs, thereby simulating the scenario. Each simulation episode concludes upon detecting RL NPC vehicle switching, collisions, or the vehicles reaching their destinations. To expedite the training process of the RL model, we establish a dynamic updating mechanism for selecting and switching attacking NPC vehicles. We prioritize the vehicle within a certain distance ahead of the ego vehicle that has a lower speed and presents collision risks for control by the RL model. In the absence of such vehicles, we select the vehicle nearest to the ego vehicle for RL control. Additionally, vehicles under RL control are periodically switched according to fixed time intervals, adhering to the aforementioned selection criteria. This setup ensures that in long-term scenarios, attacking vehicles can continuously engage with the ego vehicle, allowing the RL model to undergo sustained learning. The states of the ego vehicle, RL NPC, and NDD vehicles—including position, Lane ID, speed, and other relevant data—are collected and input into the RL model. Based on these collected states, the RL model outputs the action for the attacking NPC vehicle. Subsequently, a three-layer relative safety criticality reward is calculated. Throughout the entire process, the state $s_t$, action $a_t$, reward $r_t$ at step $t$, and the state $s_{t+1}$ at step $t+1$ in one episode are combined into an experience buffer. This buffer is stored in memory for updating the parameters of the RL model.

In more detail, the AuthSim framework treats the generation of safety-critical scenarios as an optimization problem, modeled as follows
\begin{equation}
\centering
\begin{split}
  & \min_{\theta} ~-J(\theta), \\
  &\ \text{s.t.} ~\theta \in \bm{\Theta},
\end{split}
\label{equ:problem_def}
\end{equation}
where $J(\theta)$ is the objective of scenario criticality defined in Equation \ref{equ:safety_criticality}, and $\theta$ represents the parameters to be optimized and $\bm{\Theta}$ encompasses all reasonable values that $\theta$ can assume.




The safety-critical scenario generation described in Equation \ref{equ:problem_def} constitutes a black-box optimization problem. Consequently, the optimization process cannot access the derivatives of the scenario criticality objective function. To overcome this challenge, we employ reinforcement learning for safety-critical scenario generation. This approach enables decision-making by maximizing cumulative rewards through interactions with the environment.


Here we model the process of NPC vehicles attacking the ego vehicle as a Markov Decision Process (MDP). The MDP is defined as $M = \langle S, A, P, R, \gamma \rangle$, where $S$ represents the set of observed states of NPC and ego vehicles, $A$ denotes the set of possible actions the attacking NPC vehicle can take, $P$ is the transition probability based on the NPC's action, $R$ is the reward function, and $\gamma$ is the discount factor.


\begin{itemize}
\item \textbf{State.} To guide the NPC vehicle in taking rational adversarial actions against the ego vehicle, it is crucial that the NPC can perceive the distance and speed relationships between itself and the ego vehicle. Empirically, the danger, boundary, and safety distances of a vehicle in the y-axis (lateral) direction are typically minimal. For simplicity, we set these distances as fixed small values and focus on modeling the three-layer danger, boundary, and safety distances along the x-axis (longitudinal). The observed states consist of nine components: the relative danger distance $d_{x,danger}$ defined in Equation \ref{equ:x_danger}, the relative boundary distance $d_{x,boundary}$ defined in Equation \ref{equ:x_boundary}, the relative safety distance $d_{x,safety}$ defined in Equation \ref{equ:x_safe}, the NPC's position $(x_{NPC},y_{NPC})$, the indicator $i$ to identify whether the NPC and the ego vehicle are in the same lane, the $x-$axis and $y-$axis distances $d_{x,ego,NPC}$, $d_{y,ego,NPC}$ between the NPC vehicle and the ego vehicle, the relative distance $d_{rel}$ and velocity $v_{rel}$ between the NPC and the ego vehicle.


\item \textbf{Action.} To expedite the RL model training, we have constrained the action space of the attacking NPC vehicle. The attacking NPC vehicle controlled by the RL model can take five actions: left turn, right turn, decelerate with maximum acceleration, maintain a constant speed, and accelerate with maximum acceleration.



\item \textbf{Reward.} To solve the optimization problem define in Equation \ref{equ:problem_def}, we calculate the reward in time step $t$ as
\begin{equation}
\centering
\begin{split}
  r_{t} = &\sum_{i=1}^N\bigg( P_{danger}S_{danger}^{i,t}\\
&+P_{boundary}S_{boundary}^{i,t}+P_{safety}S_{safety}^{i,t} \bigg).
\end{split}
\label{equ:reward}
\end{equation}



\end{itemize}

\section{Experiment results}
\label{sec:exp_results}

The proposed AuthSim framework for scenario generation is extensively evaluated to validate its effectiveness in enhancing the efficiency and authenticity of scenario generation. Comprehensive experiments are conducted using the Highway-env simulator to assess the performance of the proposed approach. To ensure a consistent benchmark, the autonomous driving system tested in this research employs the same deep reinforcement learning (DRL) pre-trained model as \cite{feng2021intelligent} for comparison. The specific configuration parameters of the experiments are detailed in Table \ref{tab:exp_para}.

\begin{table}[!htpb]
    \centering
\begin{tabular}{c|c}
\hline  
Simulator &Highway-env \\
\hline
$a_{x,ego,max,dece}$ & $4~m/s^2$\\
\hline
$a_{x,ego,max,accel}$ & $2~m/s^2$\\
\hline
$a_{x,ego,min,dece}$ & $0.2~m/s^2$\\
\hline
$\rho$ & $0.3~s$\\
\hline
\end{tabular}
\caption{Configuration parameters of our experiments.}
\label{tab:exp_para}
\end{table}

The RL agent for safety-critical scenario generation is trained on a 3-lane highway over 20000 episodes. In each episode, the positions and speeds of the NPC vehicles are randomly initialized. The ego vehicle drives a fixed distance, and at set intervals, the attacking NPC vehicle controlled by the RL agent is chosen from the NPC vehicles within a 75-meter range of the ego vehicle. Other NPC vehicles follow the driving patterns described in \cite{feng2021intelligent}, adhering to the naturalistic driving data (NDD).

We compare our AuthSim framework with both non-RL methods (NADE) and RL methods (TTC, TTB, and DRAC). The NADE method builds on the NDD method by randomly selecting the vehicle around the ego vehicle and increasing the likelihood of the vehicle attacking the ego vehicle through importance sampling, thereby enhancing the efficiency of generating critical test scenarios. The RL methods, including TTC, TTB, and DRAC, follow the same training and testing process as AuthSim to generate critical test scenarios, differing only in the RL reward function. By comparing our proposed AuthSim framework with these benchmark methods, we can verify its advantages in generating safety-critical autonomous driving test scenarios.

\subsection{Efficiency analysis}

\begin{figure*}[!htpb]
	\centering
\includegraphics[width=1.0\textwidth]{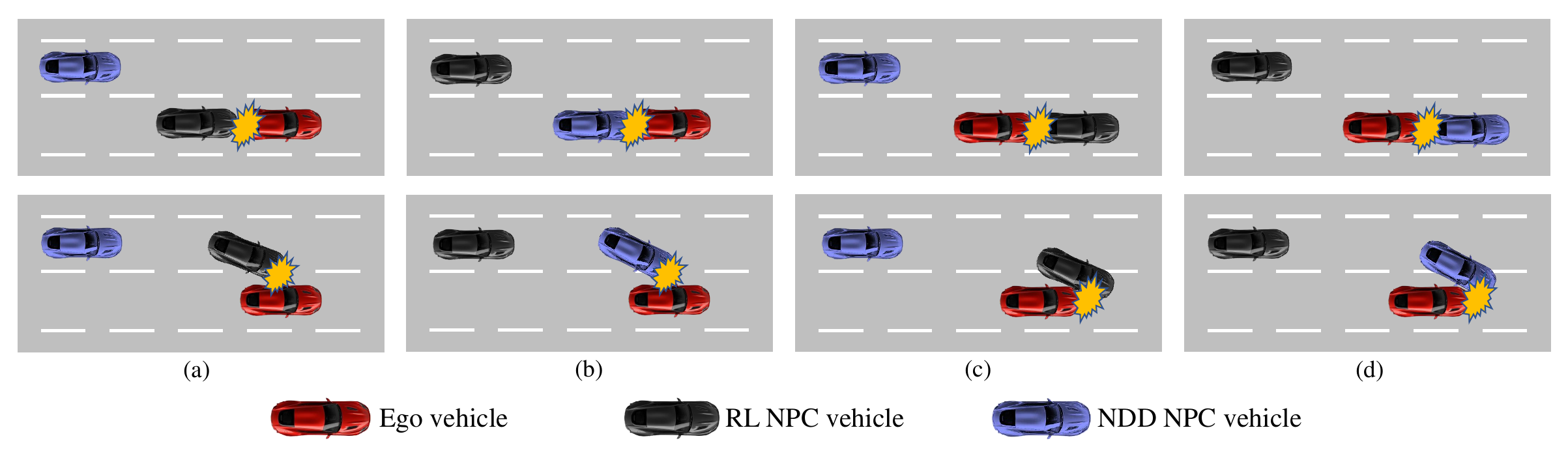} 
	\caption{Categories of 4-type collisions. (a) The RL NPC vehicle collides with the ego vehicle. (b) The NDD NPC vehicle collides with the ego vehicle. (c) The ego vehicle collides with the RL NPC vehicle. (d) The ego vehicle collides with the NDD NPC vehicle.}
	\label{fig:collision-cat4}
\end{figure*}

\begin{table*}[!htpb]
    \centering
    \resizebox{1.0\textwidth}{!}{\begin{tabular}{cccccccccc}
\hline  
Method & RL$\rightarrow$ego & NDD$\rightarrow$ego &ego$\rightarrow$RL &ego$\rightarrow$NDD  &Invalid collisions& Valid collisions &All collisions & Valid collisions/All collisions(\%)  &Valid collision/All tests(\%)\\
\hline 
NADE & /   & 76 & /  & 116  & 76 & 116 & 192 & 60.4\% & 11.6\% \\
TTC  & 49 & 19 & 206 & 16   & 68 & 222  & 290 & 76.6\% & 22.2\% \\
TTB  & 72 & 22 & 153 & 9    & 94 & 162  & 256 & 63.3\% & 16.2\% \\
DRAC & 69 & 16  & 184 & 21  & 85 & 205 & 290 & 70.7\% & 20.5\%\\
\textbf{AuthSim (ours)} &8 &9 & 254 & 16  & 17 & \textbf{270} & 287&\textbf{94.1\%} &\textbf{27.0\%} \\
\hline
\end{tabular}}
\caption{Scenario efficiency comparison of different methods.}
\label{tab:crash_type}
\end{table*}

The purpose of testing autonomous driving vehicles is to identify and correct their deficiencies in various scenarios. During testing, numerous scenarios arise that pose threats to the ego vehicle and result in collisions. However, not all collision scenarios are meaningful for evaluating autonomous driving. Many incidents involve the ego vehicle driving normally while the NPC vehicle intentionally causes a collision. Examples include the NPC vehicle rear-ending the ego vehicle without sudden braking by the ego vehicle or intentionally hitting the side of the ego vehicle when it is driving normally. These collisions are primarily the responsibility of the NPC vehicle and are unrelated to the ego vehicle's driving behavior. Such scenarios have little significance for testing the ego vehicle and do not contribute to improving its strategies.

Methods, like NADE, TTC, TTB, and DRAC, do not restrict the NPC vehicle's methods of attacking the ego vehicle, often encouraging the NPC vehicle to collide with the ego vehicle in various ways. Consequently, many scenarios involve the NPC vehicle intentionally crashing into the ego vehicle, which are not useful for improving the ego vehicle's driving strategies. Although these scenarios involve collisions, they hold no practical significance for testing. We collectively refer to these as meaningless testing scenarios. 

The AuthSim framework encourages NPC vehicles to enter the safe boundary area of the ego vehicle by establishing a three-layer relative safety region model and setting relevant reward functions in the reinforcement learning model. This approach reduces the frequency of intentional rear-end and side collisions with the ego vehicle, generating more meaningful testing scenarios for evaluating ego vehicles.

As described in Figure \ref{fig:collision-cat4}, we classified the collision scenarios into four categories: (1) the NPC vehicle controlled by RL actively collides with the ego vehicle (represented by $RL \rightarrow ego$), (2) the NPC vehicle controlled by NDD actively collides with the ego vehicle (represented by $NDD \rightarrow ego$), (3) the ego vehicle collides with the NPC vehicle controlled by RL (represented by $ego \rightarrow RL$), and (4) the ego vehicle collides with the NPC vehicle controlled by NDD (represented by $ego \rightarrow NDD$). The first two are considered invalid collision scenarios, while the latter two are considered valid collision scenarios. We conducted statistical analyses on collision scenarios using the methods of NADE, TTC, TTB, and DRAC, respectively. 

For each method, we generate a total of 1,000 scenarios. The results present in Table \ref{tab:crash_type} demonstrate that, while the total number of collision scenarios generated by our framework is not the highest (with 287 collision scenarios, compared to 290 generated by TTC and DRAC), our framework yields the highest number of effective collision scenarios (with 270 effective collisions, compared to 222 and 205 generated by TTC and DRAC, respectively). The proposed AuthSim framework encounters a total of 287 collisions, of which 270 are effective collision scenarios. This results in a frequency of 27\% for generating effective collision scenarios and an 94.1\% ratio of effective collision scenarios to total collisions. Among all evaluated methods, the TTC approach obtains the closest result to our AuthSim framework in terms of efficiency for generating safety-critical scenarios. Its effective collision scenario generation frequency stands at 22.2\%, and the proportion of effective collision scenarios to total collisions is 76.6\%. These values are 4.8\% and 17.5\% lower, respectively, compared to the performance of our proposed AuthSim framework. Other methods, including NADE, TTB, and DRAC, exhibit significant differences from our AuthSim Framework in terms of the frequency of generating effective collision scenarios and the proportion of effective collision scenarios to total collisions. 

Overall, the proposed AuthSim framework has effectively guided the attack behavior of NPC vehicles by establishing the three-layer relative safety region model, thereby reducing the incidence of irrelevant scenarios in autonomous driving tests. Our approach has demonstrated a higher efficiency in generating effective safety-critical scenarios compared to existing methods. Additionally, we have significantly improved the ratio of effective collision scenarios to total collision scenarios. These enhancements enable the generation of more authentic safety-critical scenarios within the same testing cycles and substantially reduce the time required to screen out invalid collision scenarios.

\subsection{Authenticity validation}

Our objective is to eschew scenarios that are extreme, contrived, and devoid of practical significance in the realm of autonomous driving. Instead, we aim to cultivate simulations that are more authentic, natural, and reflective of the true capabilities and limitations of autonomous driving vehicles. To analyze the authenticity of safety-critical scenarios generated by the AuthSim framework, we focus on NPC lane-changing safety-critical scenarios. This focus is driven by two main reasons. First, NPC lane-changing scenarios constitute the highest proportion ($\textgreater 75\%$) of safety-critical scenarios generated by all methods, as shown in Table \ref{tab:exp_auth}. Second, determining the time for NPC vehicles to initiate attacks and the time left for ego vehicles to react in various other types of scenarios present significant challenges. However, in lane-changing scenarios, we can utilize NPC vehicles crossing lane lines as the trigger for launching attacks, facilitating an effective assessment of the rationality of collision scenarios.

\begin{table}[!htpb]
    \centering
    \resizebox{0.45\textwidth}{!}{\begin{tabular}{cccc}
\hline  
Method &NPC lane-changing &Valid collision &Ratio\\
\hline  
NADE & 88 &116 & 75.8\% \\
TTC & 180 &212 & 84.9\%\\
TTB & 121 & 155 & 78.1\% \\
DRAC  &  162 &200 & 81.0\%\\
Authsim & 232 &270 & 85.9\%\\
\hline
\end{tabular}}
\caption{The ratios of NPC lane-changing scenarios in effective collision scenarios.}
\label{tab:exp_auth}
\end{table}

\begin{figure}[!htpb]
	\centering
\includegraphics[width=0.4\textwidth]{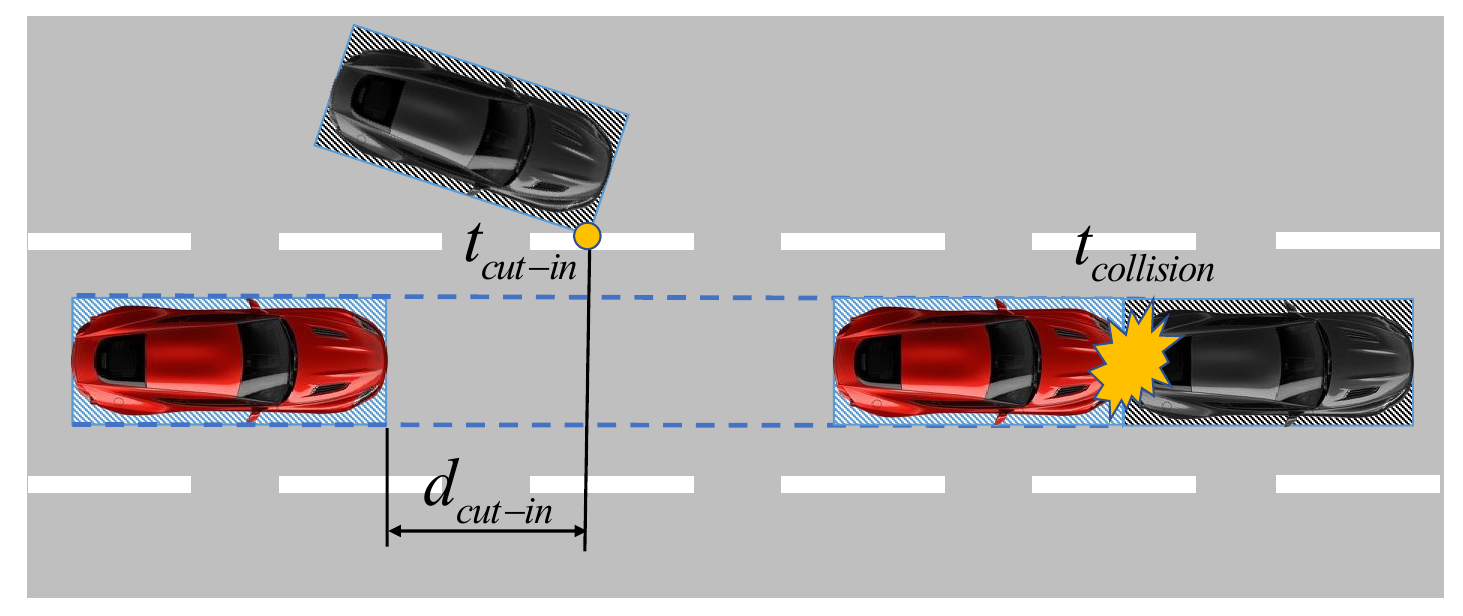} 
	\caption{Evaluation of scenario authenticity.}
	\label{fig:eval}
\end{figure}


Here, empirically, we introduce two intuitive indicators $d_{cut-in}$ and $t_{interval}$ to evaluate the authenticity of the scenarios. In the collision scenario depicted in Figure \ref{fig:eval}, $t_{cut-in}$ denotes the moment when the vehicle initiates a lane change maneuver and makes contact with the lane markings, while $d_{cut-in}$ indicates the distance between the NPC and the ego vehicles at the time $t_{cut-in}$. $t_{collision}$ represents the exact moment when the collision occurs. Here, the time interval $t_{interval}$ between the beginning of cutting-in and collision is

\begin{equation}
\centering
t_{interval}=t_{collision}-t_{cut-in}.
\label{equ:t_inter}
\end{equation}

Evidently, a smaller value of $t_{interval}$ or $d_{cut-in}$ indicates a higher degree of scenario extremeness and deliberateness, while simultaneously lowering the authenticity and naturalness of the scenario. Consequently, both the cutting-in distance, $d_{cut-in}$, and the time interval, $t_{interval}$, can serve as valuable indicators for evaluating the authenticity and naturalness of the safety-critical scenarios.

\begin{figure*}
	\centering
	\subfigure[]{\includegraphics[width=0.3\textwidth]{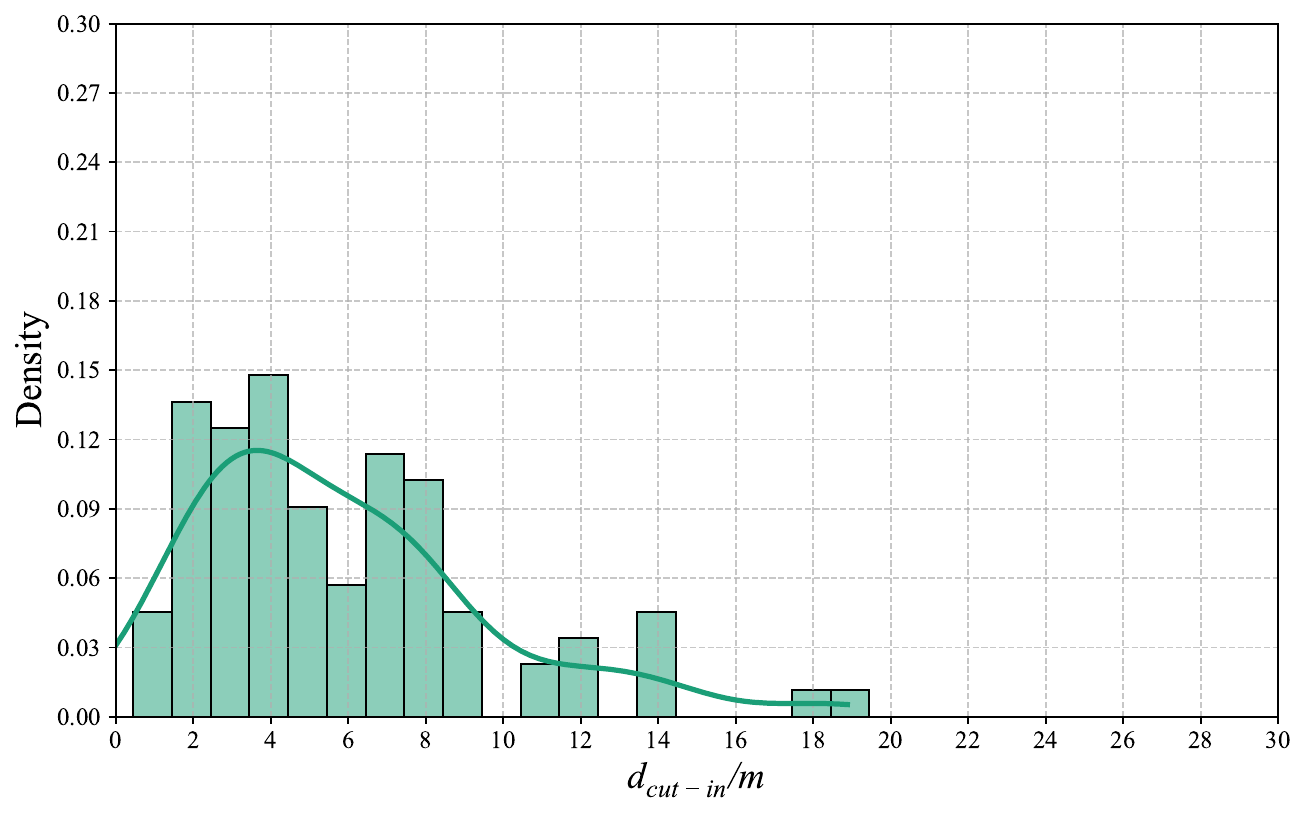} \label{fig:d_NADE}}
	\subfigure[]{\includegraphics[width=0.3\textwidth]{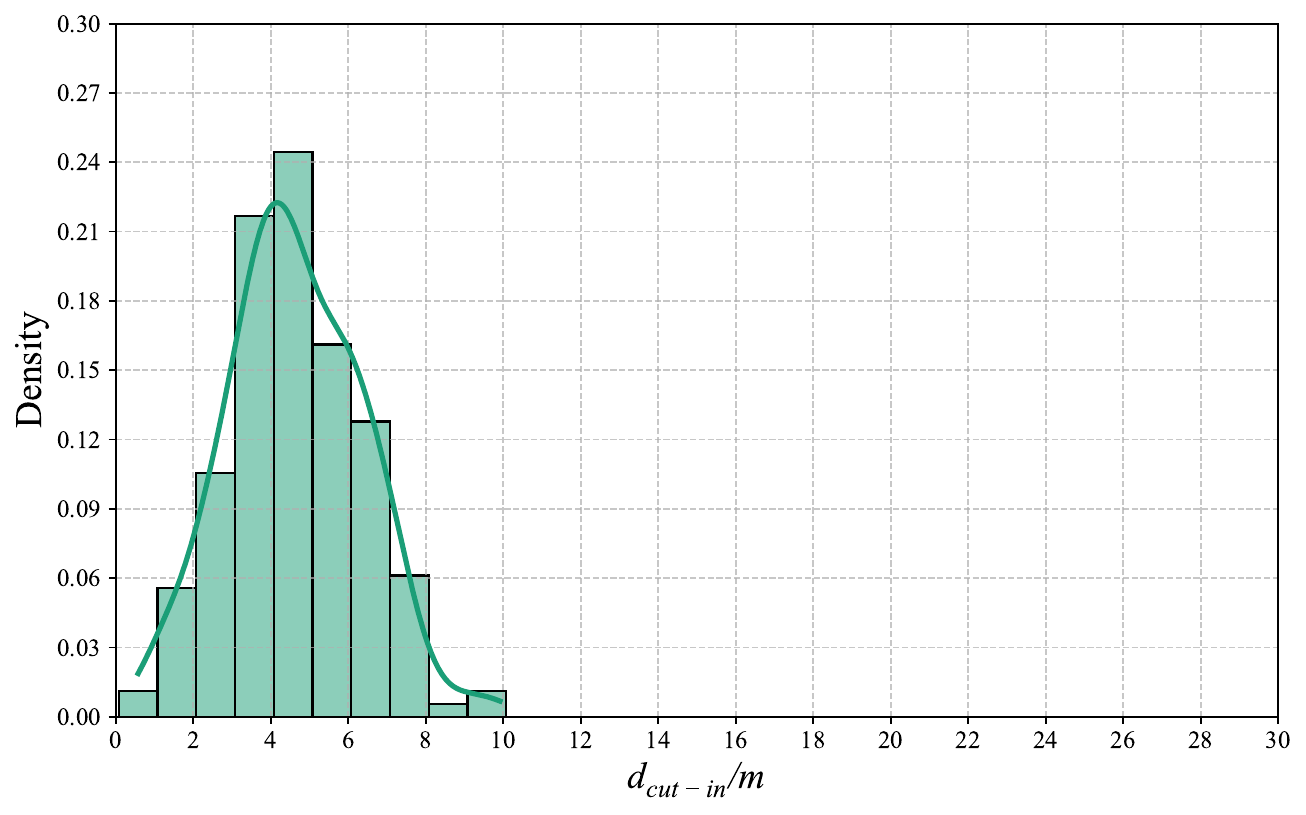} \label{fig:d_TTC}}
	\subfigure[]{\includegraphics[width=0.3\textwidth]{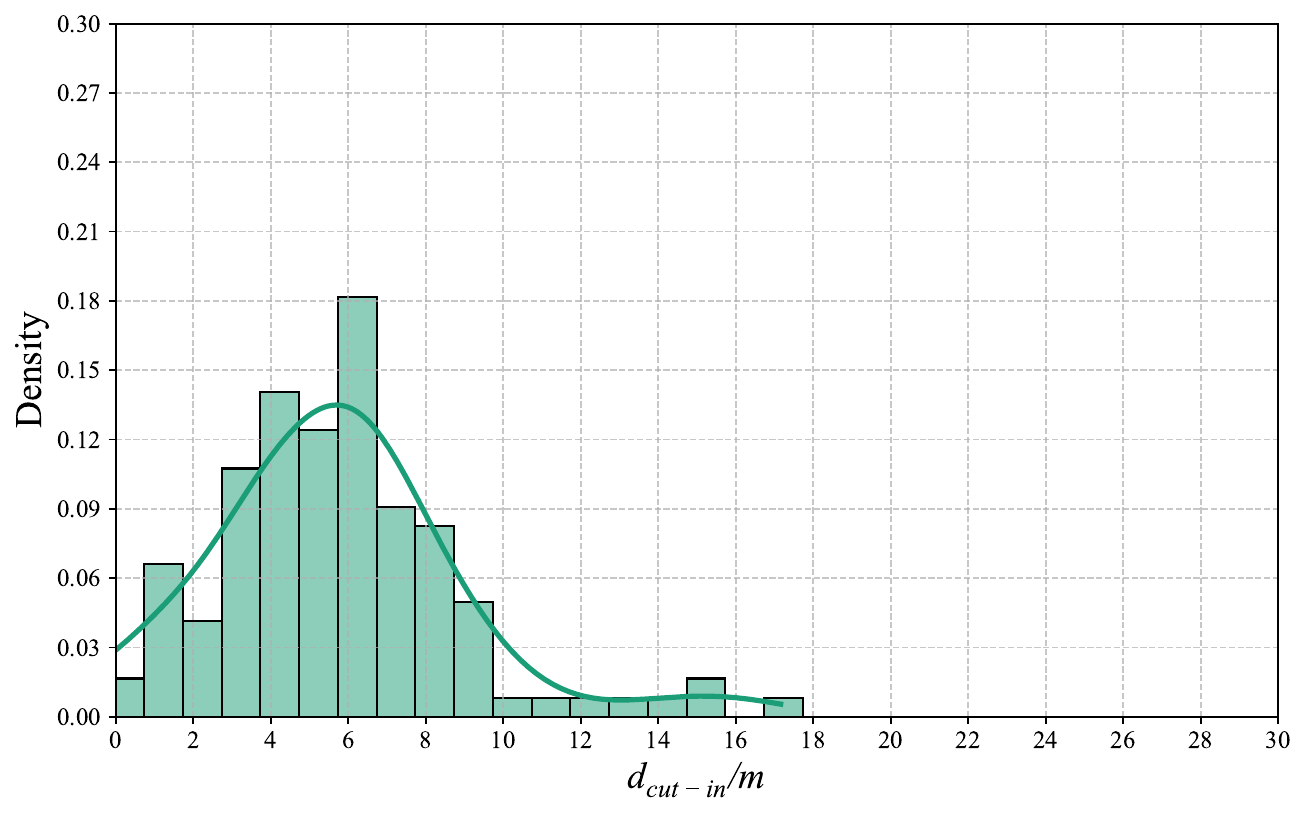} \label{fig:d_TTB}}
	\subfigure[]{\includegraphics[width=0.3\textwidth]{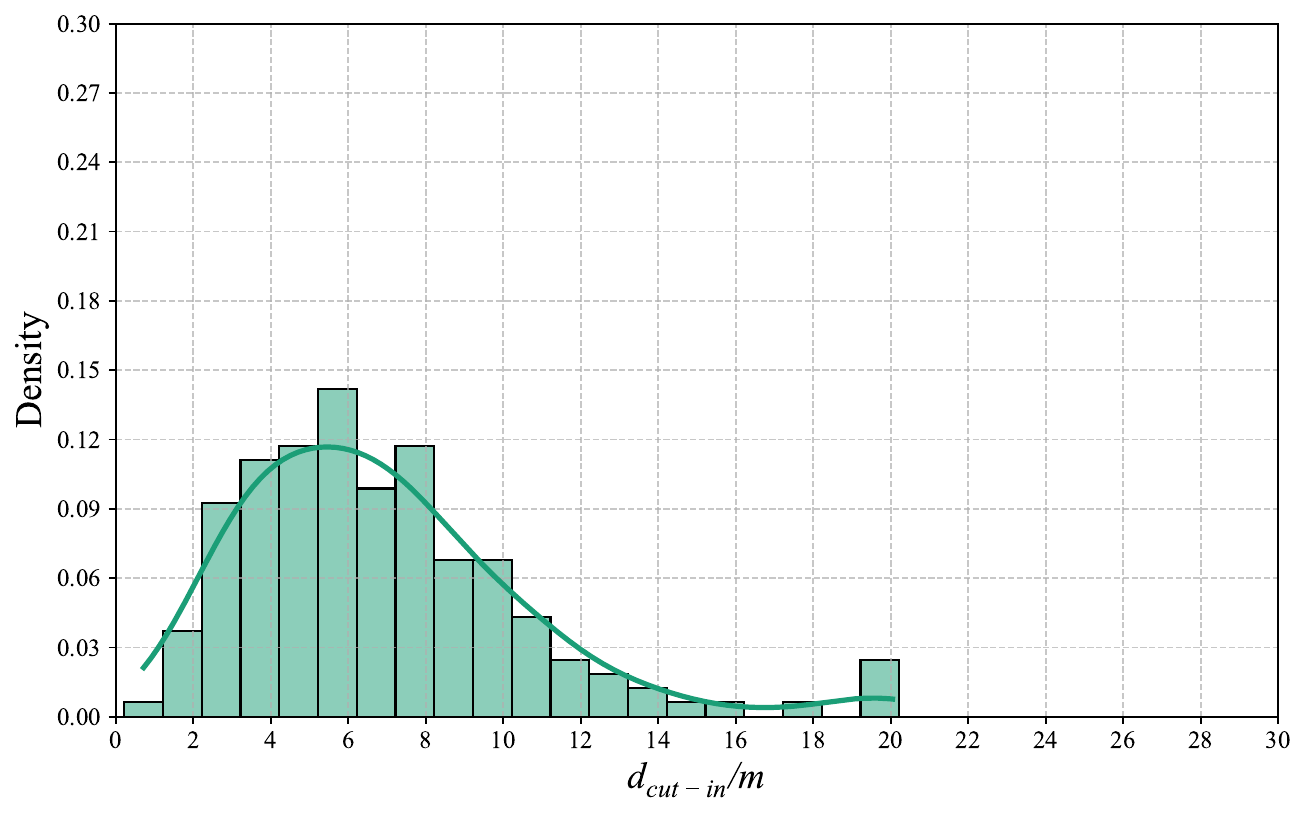} \label{fig:d_DRAC}}
     \subfigure[]{\includegraphics[width=0.3\textwidth]{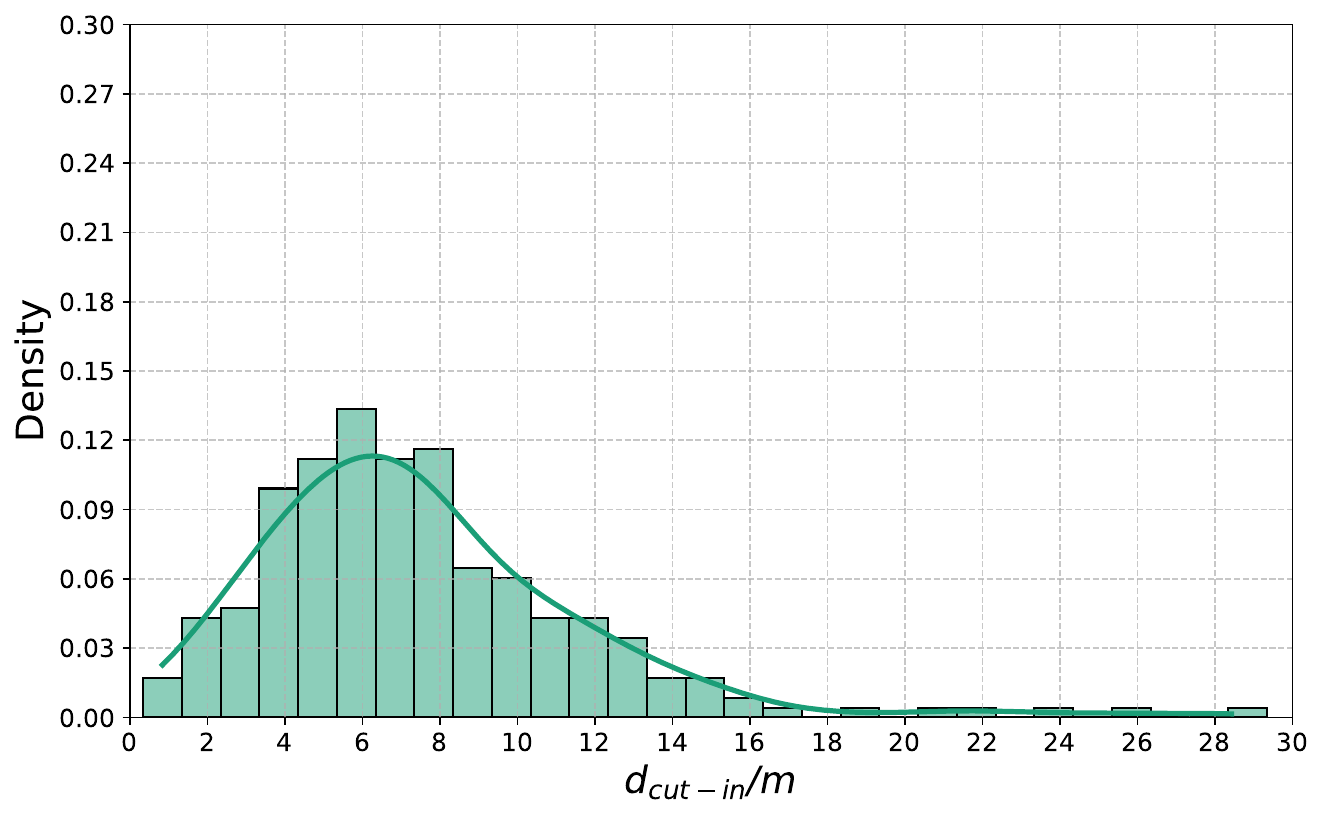} \label{fig:d_auth}}
	\caption{Distributions of the cut-in distance of safety-critical scenarios generated by different methods. (a) NADE; (b) TTC; (c) TTB; (d) DRAC; (e) AuthSim.}
	\label{fig:d_result}
\end{figure*}

\begin{figure*}
	\centering
	\subfigure[]{\includegraphics[width=0.3\textwidth]{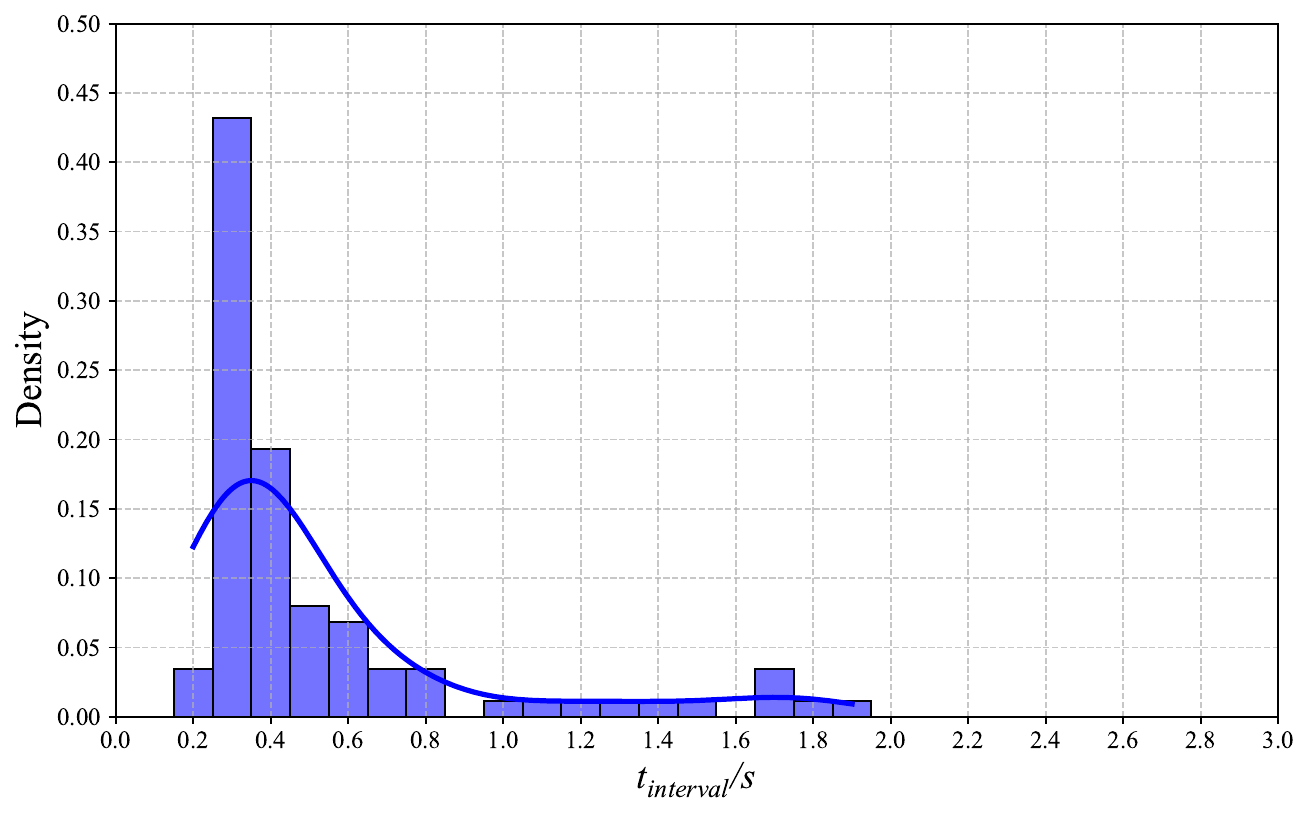} \label{fig:t_NADE}}
	\subfigure[]{\includegraphics[width=0.3\textwidth]{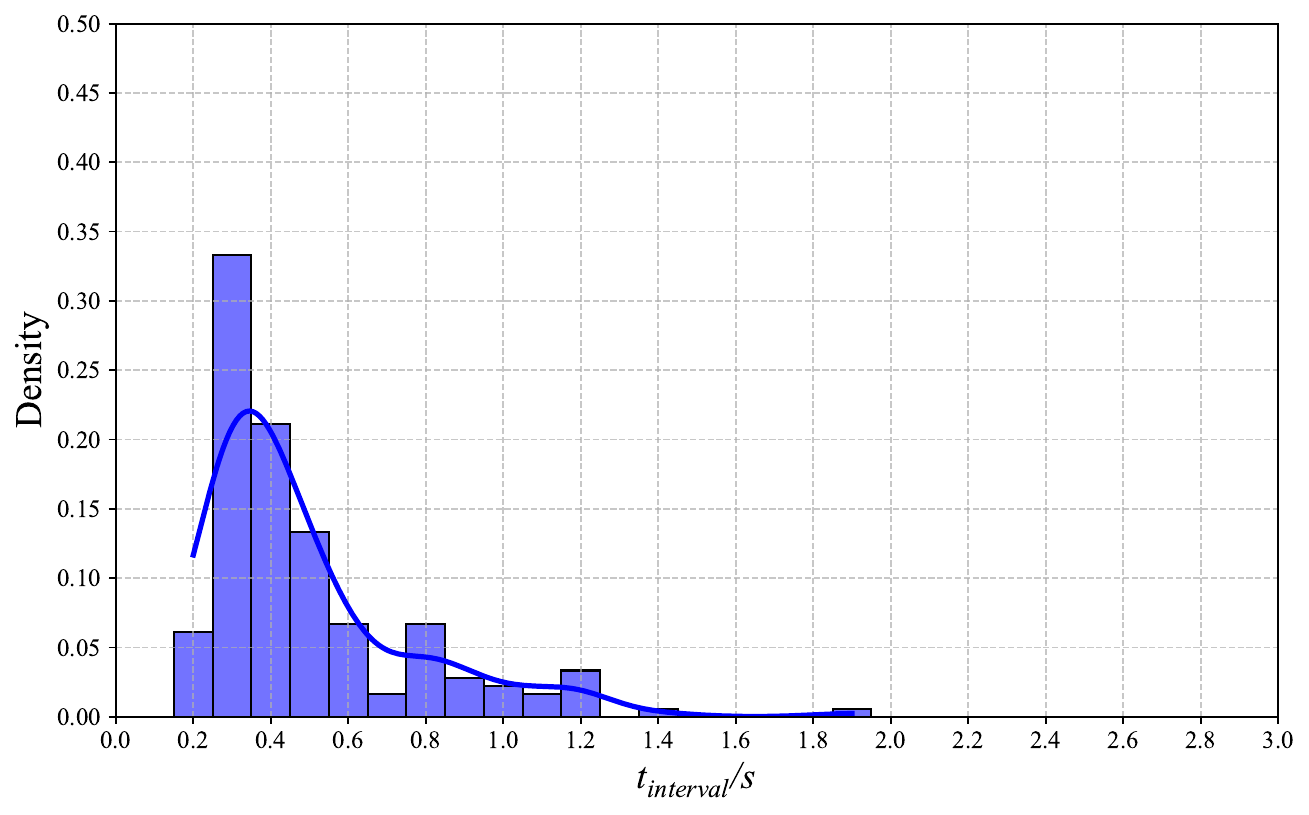} \label{fig:t_TTC}}
	\subfigure[]{\includegraphics[width=0.3\textwidth]{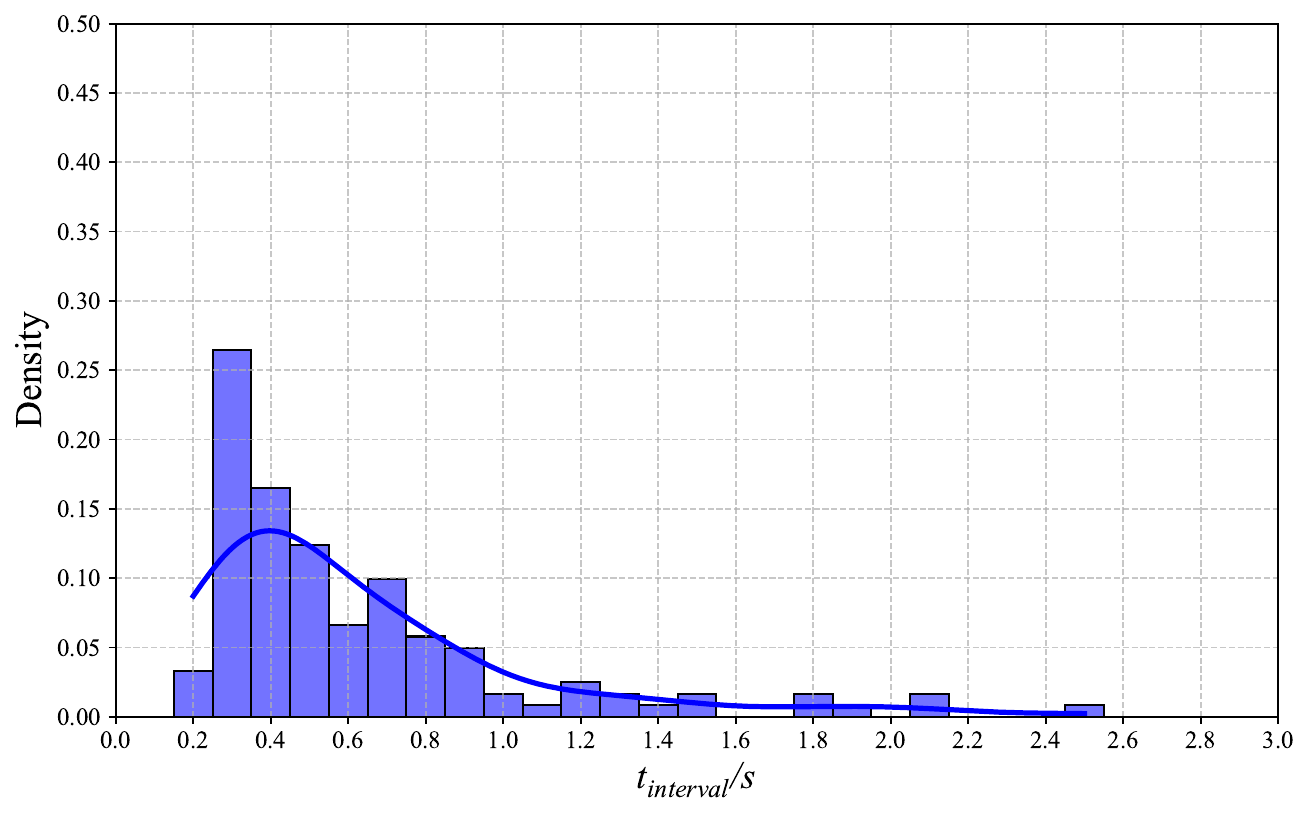} \label{fig:t_TTB}}
	\subfigure[]{\includegraphics[width=0.3\textwidth]{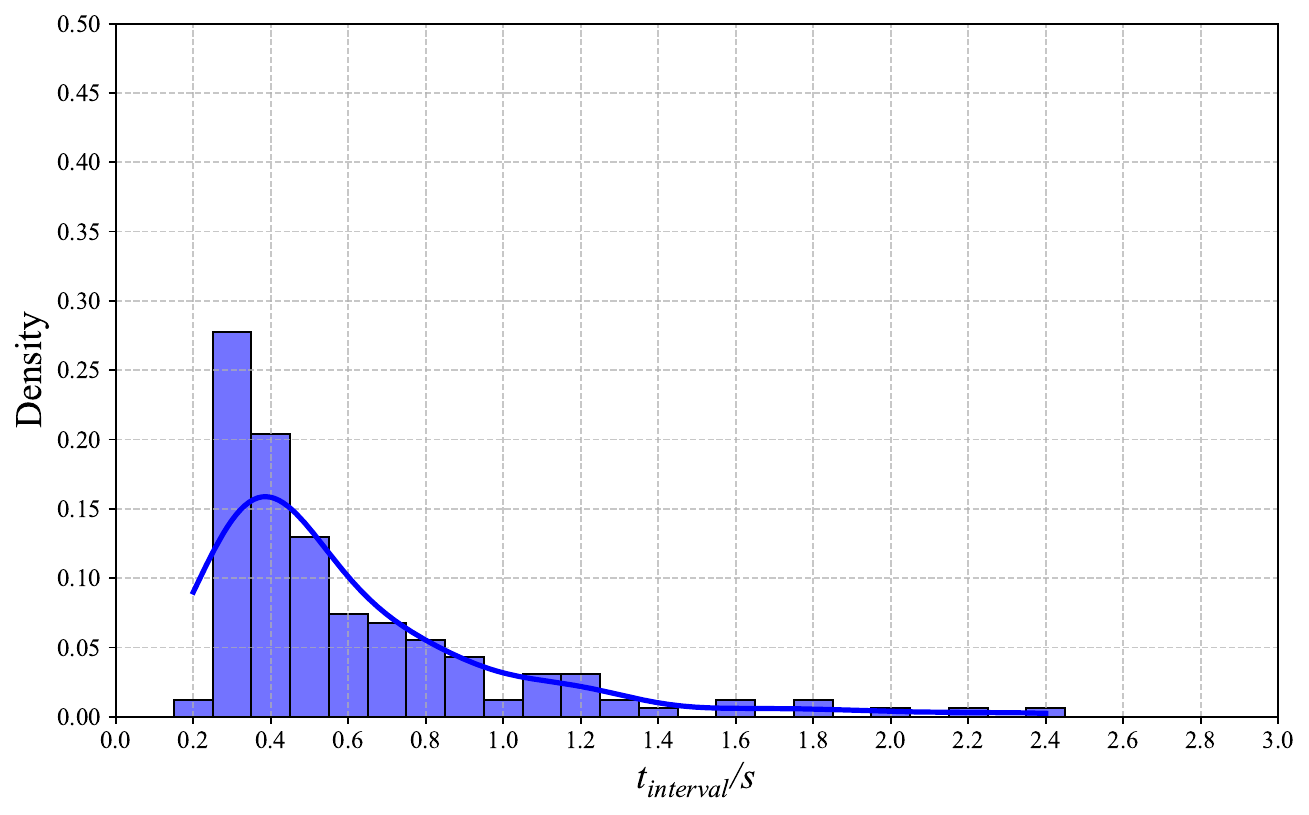} \label{fig:t_DRAC}}
     \subfigure[]{\includegraphics[width=0.3\textwidth]{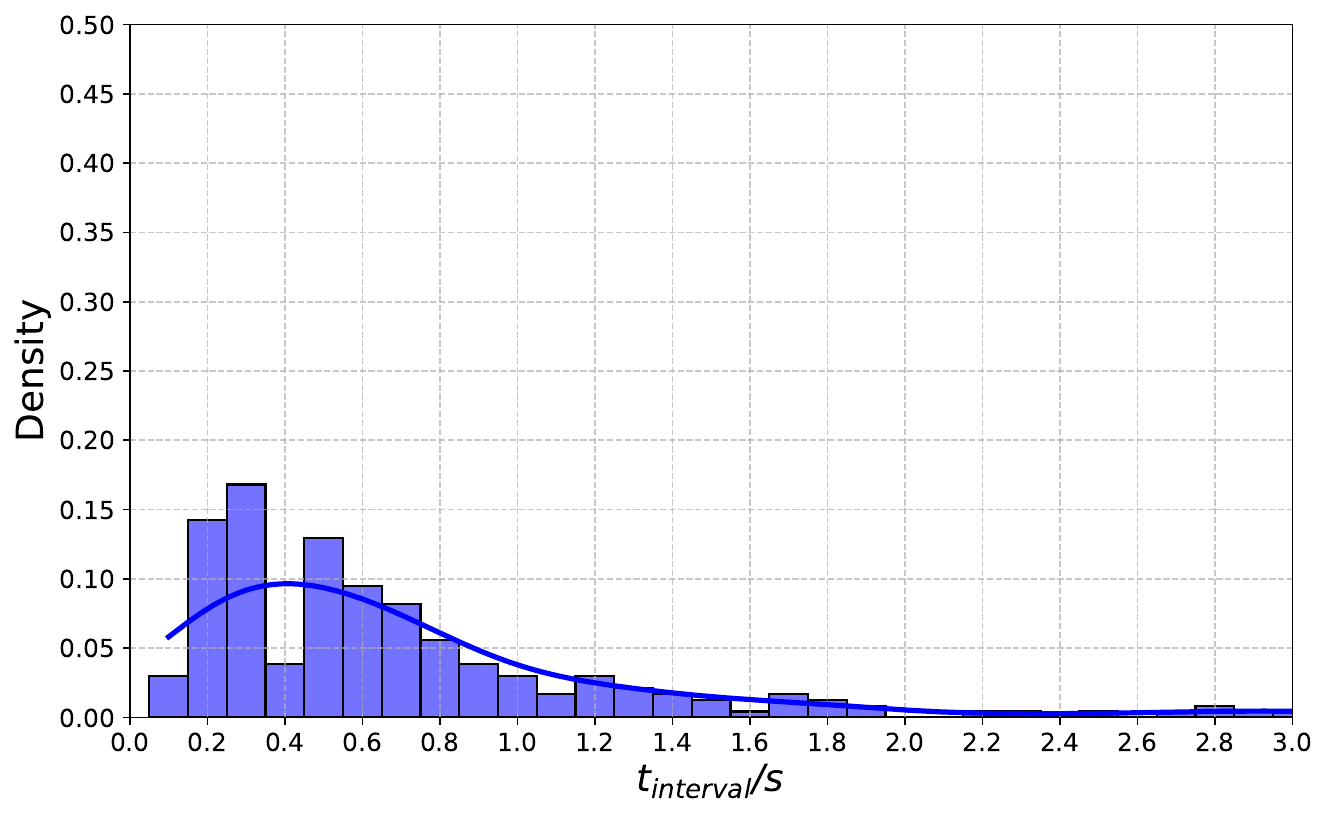} \label{fig:t_auth}}
	\caption{Distributions of the collision time interval of safety-critical scenarios generated by different methods. (a) NADE; (b) TTC; (c) TTB; (d) DRAC; (e) AuthSim.}
	\label{fig:t_result}
\end{figure*}

\begin{table}[!htpb]
    \centering
    \resizebox{0.48\textwidth}{!}{
    \begin{tabular}{ccc}
\hline  
Method &$d_{cut-in}(m)$/Improvement  &$t_{interval}(s)$/Improvement \\
\hline  
NADE & 5.73/23.76\%  & 0.53/8.16\% \\
TTC (Baseline)  & 4.63/- & 0.49/- \\
TTB & 5.57/20.30\% &  0.63/28.57\% \\
DRAC  & 6.86/48.16\%  & 0.59/20.41\%\\
\textbf{Authsim (ours)} &\textbf{7.22/55.94\%} &\textbf{0.75/53.06\%} \\
\hline
\end{tabular}
}
\caption{Scenario authenticity comparisons of different methods.}
\label{tab:exp_lane_change}
\end{table}

We collect and analyze NPC lane-changing collision scenario data generated by various methods, including NADE, TTC, TTB, DRAC, and our proposed AuthSim. We calculate the mean values of the cut-in distance ($d_{cut-in}$) and the collision time interval ($t_{interval}$), with the results presented in Table \ref{tab:exp_lane_change}. AuthSim achieves the best results in our study. Using the most commonly employed TTC indicator as a baseline, we observe a 55.94\% improvement in average cut-in distance and a 53.06\% improvement in average collision interval time. Compared to other methods like NADE, TTB, and DRAC, our method also shows significant enhancements. For instance, when compared to the top-performing DRAC method, our method demonstrates improvements of 5.25\% in the average cut-in distance and 27.12\% in the average collision interval time. It is worth noting that the 27.12\% improvement in collision interval time is substantial, demonstrating the significant advantage of AuthSim in generating authentic scenarios.



The distributions of cut-in distance and collision time interval are illustrated in Figures \ref{fig:d_result} and \ref{fig:t_result}, respectively. Regarding cut-in distance, compared to NADE, TTC, and TTB, AuthSim significantly reduces the probability of NPC vehicles cutting in at short distances and decreases the number of extreme cut-in scenarios. Additionally, compared to DRAC, our approach also reduces the probability of NPC vehicles cutting in at shorter distances and demonstrates a favorable distribution of long cut-in distances for NPC vehicles. Our advantage is particularly evident in terms of collision time intervals. As shown in Figure \ref{fig:t_result}, our method significantly reduces the distribution of collision intervals in lower-value areas compared to other methods. Furthermore, our approach results in longer collision intervals.

In summary, compared to existing methods, AuthSim achieves significantly higher cut-in distances and longer collision time intervals. This indicates that when NPC vehicles initiate attacks on ego vehicles, the ego vehicles have more time to react, thereby significantly reducing the extremity of the attacks. This enhancement contributes to generating safety-critical scenarios more authentically and naturally.

\section{Conclusion}
\label{sec:conclusion}

Scenario-based autonomous driving tests are crucial for ensuring vehicle safety. Traditional evaluation metrics, like TTC, do not restrict NPC vehicles, leading to extreme and deliberate collision scenarios. These scenarios often result in unavoidable collisions attributed to NPC vehicles, offering little testing significance for autonomous vehicles. To address this, we propose a three-layer relative safety region model for evaluating scenario criticality. This model guides NPC vehicles to enter more ego vehicle safety boundaries, improving the rationality and authenticity of safety-critical scenarios. We have developed a framework called AuthSim, combining the three-layer relative safety region model and reinforcement learning, and conducted extensive comparisons with classical methods such as NADE, TTC, TTB, and DRAC. Experimental results show that our method significantly increases cut-in distance and collision interval time for NPC vehicles while maintaining higher efficiency in generating effective safety-critical scenarios. This provides ego vehicles with more reaction time and reduces the frequency of extreme collision scenarios. In summary, our method offers a more effective adversarial scenario generation approach for autonomous driving tests, enhancing safety development. Looking forward, we aim to further improve the complexity and optimization efficiency of autonomous driving adversarial scenarios to accelerate the application of autonomous driving.

\bibliographystyle{unsrt} 
\bibliography{ref}
\end{document}